\documentclass{article}
\usepackage{arxiv}

\usepackage[utf8]{inputenc} 
\usepackage[T1]{fontenc}    
\usepackage{hyperref}       
\usepackage{url}            
\usepackage{booktabs}       
\usepackage{amsfonts}       
\usepackage{nicefrac}       
\usepackage{microtype}      
\usepackage{lipsum}		
\usepackage{graphicx}
\usepackage[authoryear]{natbib}
\usepackage{doi}

\usepackage{amsmath}   
\usepackage{amsfonts}  
\usepackage{amssymb}    
\usepackage{mathtools}  
\usepackage{geometry}
\usepackage{listings}
\usepackage{graphicx}
\usepackage{geometry}
\usepackage{textcomp}
\usepackage{forest}  
\usepackage{tikz} 
\usepackage{enumitem}
\usepackage{setspace}   
\usetikzlibrary{positioning} 
\usepackage{tocloft}
\usepackage{longtable}
\usepackage{tikz}
\usepackage{xstring}
\usepackage{wrapfig}
\usepackage{booktabs}
\usepackage{array}
\usepackage{float}
\usepackage{listings}
\usepackage{subcaption}
\usepackage{pdflscape}
\usepackage{pgfplots}
\usepackage{multirow}
\usepackage{multicol}
\usepackage{ragged2e}
\pgfplotsset{compat=1.18}
\usepgfplotslibrary{fillbetween}

\definecolor{darkblue}{RGB}{0,0,139}
\hypersetup{colorlinks=true, citecolor=darkblue, linkcolor=darkblue, urlcolor=darkblue}
\usetikzlibrary{shapes,arrows,positioning}

\tikzstyle{block} = [rectangle, draw, fill=blue!20, 
    text width=3em, text centered, rounded corners, minimum height=3em]
\tikzstyle{line} = [draw, -latex']

\title{Evaluation of Finetuned Large Language Models \\ in Abstract Meaning Representation Parsing}

\date{August, 2025}	

\author{Ho Shu Han \\
	Department of Computer Science\\
	University College London\\
	66-72 Gower St, London WC1E 6EA, United Kingdom \\
	\texttt{zcabshh@ucl.ac.uk}
}

\renewcommand{\headeright}{Evaluation of Finetuned LLMs in AMR Parsing}
\renewcommand{\undertitle}{}
\renewcommand{\shorttitle}{}

\hypersetup{
pdftitle={Evaluation of Finetuned LLMs in AMR Parsing},
pdfsubject={LLM, AMR Parsing},
pdfauthor={Shu Han Ho},
pdfkeywords={LLM, AMR Parsing},
}
\begin{document}
\maketitle

\section{Abstract}
Abstract Meaning Representation (AMR) \citep{banarescu-etal-2013-abstract} is a semantic formalism that encodes sentence meaning as rooted, directed, acyclic graphs \citep{lyu-etal-2021-differentiable}, where nodes represent concepts and edges denote semantic relations \citep{sameAMR}. 

Finetuning decoder-only Large Language Models (LLMs) represent a promising novel straightfoward direction for AMR parsing. 
This paper presents a comprehensive evaluation of finetuning four distinct LLM architectures—Phi-3.5 \citep{abdin2024phi3technicalreporthighly}, Gemma-2 \citep{gemmateam2024gemma2improvingopen}, LLaMA-3.2 \citep{githubLLaMA3MODEL_CARDmdMain}, and DeepSeek-R1-LLaMA-Distilled \citep{deepseekai2025deepseekr1incentivizingreasoningcapability} using the LDC2020T02 Gold AMR3.0 test set \citep{LDC2020T02}.

Our results have shown that straightfoward finetuning of decoder-only LLMs can achieve comparable performance to complex State-of-the-Art (SOTA) AMR parsers. 
Notably, LLaMA-3.2 demonstrates competitive performance against SOTA AMR parsers given a straightforward finetuning approach. We achieved SMATCH F1: 0.804 on the full LDC2020T02 test split, on par with APT + Silver (IBM) at 0.804 \citep{zhou2021amr} and approaching Graphene Smatch (MBSE) at 0.854 \citep{lee-etal-2022-maximum}. 
Across our analysis, we also observed a consistent pattern where LLaMA-3.2 leads in semantic performance while Phi-3.5 excels in structural validity. 

\newpage
\section{Introduction}
A core challenge in Natural Language Processing (NLP) is to move beyond surface-level text and accurately capture the underlying meaning of natural language. 
AMR addresses this by encoding sentence meaning as rooted, directed, acyclic graphs \citep{lyu-etal-2021-differentiable}, where nodes represent concepts and edges denote semantic relations \citep{sameAMR}, enabling models to reason about concepts and their relationships in a way that is independent of syntax. 

This capability has made AMR a foundational tool in the NLP community \citep{zabokrtsky-etal-2020-sentence, Bos2016SquibEP}, driving advances in information extraction \citep{zhang-etal-2021-fine}, machine translation \citep{song-etal-2019-semantic}, summarisation \citep{dohare2017textsummarizationusingabstract}, text generation \citep{song-etal-2016-amr}, and dialogue systems \citep{bonial-etal-2020-dialogue}. 
By enabling deeper and more accurate semantic understanding, AMR plays a crucial role in building more robust NLP systems.

\begin{figure}[h]
\centering
\begin{tikzpicture}
    \tikzstyle{box}=[draw, rounded corners, fill=blue!15, rectangle, text centered, minimum height=1cm, minimum width=3cm]
    \tikzstyle{note}=[draw, rounded corners, dotted, text width=4cm]
    
    \node[box] (input) at (1,0) {Input Sentence};
    \node[box, minimum width=4cm, minimum height=1.5cm] (llm) at (6,0) {LLM Parser};
    \node[box] (output) at (11,0) {Output AMR Graph};
    
    \draw[->] (input) -- (llm);
    \draw[->] (llm) -- (output);
    
    \node[note] at (1,2) {e.g., ``The boy wants to go.''};
    \node[note, align=left] at (11,2) {
        (w / want-01\\
        \hspace{0.5cm}:arg0 (b / boy)\\
        \hspace{0.5cm}:arg1 (g / go-01\\
        \hspace{1cm}:arg0 b))
    };
\end{tikzpicture}
\caption{AMR Parser Functionality, Example by \cite{banarescu-etal-2013-abstract}}
\label{fig:system_architecture}
\end{figure}
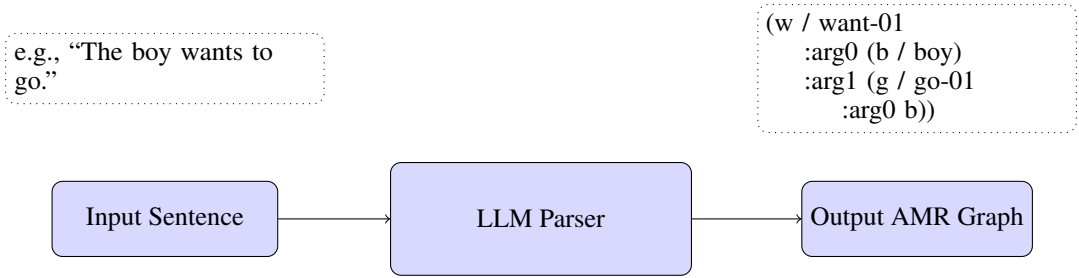

AMR parsing refers to the task of converting natural language sentences into AMRs. Figure \ref{fig:system_architecture} shows what the AMR parser should output for a given input sentence.

Traditionally, AMR parsing requires specialised architectures and complex pipelines, creating significant barriers to implementation and adaptation. 
This research demonstrates that a straightforward fine-tuning approach, using general-purpose Large Language Models (LLMs), can achieve comparable or superior accuracy while dramatically reducing implementation complexity.

By applying LoRA \citep{hu2021lora} finetuning, a Parameter-Efficient Fine-Tuning (PEFT) technique, to four distinct architectures (LLaMA-3.2, Phi-3.5, DeepSeek-R1-LLaMA-Distilled, and Gemma-2), we hope to investigate the effectiveness of LLMs in capturing semantic relationships and generating consistent graph structures.
We would also like to evaluate and compare the performance of these four models.

\subsection{Literature Review}

\subsubsection{Introduction to AMR}
\label{sec:amr_parsing}
\begin{figure}[ht]
    \centering
    \includegraphics[width=0.8\textwidth]{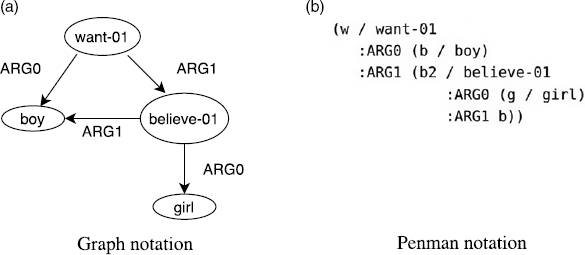}
    \caption{Example of AMR Graph and Penman Format for sentence "The boy wants the girl to believe him." \citep{turkishAMR}}
    \label{fig:amr_example}
\end{figure}

AMR \citep{banarescu-etal-2013-abstract} is a semantic formalism that encodes sentence meaning as rooted, directed, acyclic graphs \citep{zhang-etal-2019-amr}. 
This elegantly abstracts away from syntactic variations, focusing instead on capturing the underlying meaning of text, as shown in Figure \ref{fig:amr_example}.

Within this framework, there are nodes and edges \citep{Anchieta2020-zd}. 
Nodes are the concepts within the representation \citep{naseem-etal-2022-docamr}. They contain either English words (“boy”), PropBank framesets (“want-01”), or special keywords \citep{banarescu-etal-2013-abstract}. 
Keywords include special entity types (“date-entity”, “world-region”, etc.), quantities (“monetary-quantity”, “distance-quantity”, etc.), and logical conjunctions (“and”, etc) \citep{banarescu-etal-2013-abstract}.

Figure \ref{fig:amr_relations} shows the core AMR semantic relations found in nodes, grouped into categories.
It is worth noting that there are also "non-core relations, such as :beneficiary, :time, and :destination" \citep{banarescu-etal-2013-abstract}.

Edges then represent relations between these nodes, forming the semantic relationships of the representation \citep{williamson-etal-2021-intensionalizing}.

\begin{figure}[H]
    \centering
    \begin{tabular}{|p{0.23\textwidth}|p{0.67\textwidth}|}
        \hline
        \textbf{Category} & \textbf{Relations} \\
        \hline
        Frame Arguments & :arg0, :arg1, :arg2, :arg3, :arg4, :arg5 \\
        \hline
        General Semantic Relations & :accompanier, :age, :beneficiary, :cause, :compared-to, :concession, :condition, \\
        & :consist-of, :degree, :destination, :direction, :domain, :duration, \\
        & :employed-by, :example, :extent, :frequency, :instrument, :li, :location, \\
        & :manner, :medium, :mod, :mode, :name, :part, :path, :polarity, :poss, \\
        & :purpose, :source, :subevent, :subset, :time, :topic, :value \\
        \hline
        Quantity Relations & :quant, :unit, :scale \\
        \hline
        Date-Entity Relations & :day, :month, :year, :weekday, :time, :timezone, :quarter, :dayperiod, \\
        & :season, :year2, :decade, :century, :calendar, :era \\
        \hline
        List Relations & :op1, :op2, :op3, :op4, :op5, :op6, :op7, :op8, :op9, :op10 \\
        \hline
    \end{tabular}
    \caption{Categories of AMR semantic relations \citep{banarescu-etal-2013-abstract}}
    \label{fig:amr_relations}
\end{figure}
There are a few key features of AMR that are important to understand.

Firstly, different English sentences can be represented by the same AMR graph \citep{sameAMR}, as shown in Figure \ref{fig:amr_example_syntax}. 
This is done by using PropBank framesets to abstract away from English syntax \citep{banarescu-etal-2013-abstract}. 
Figure \ref{fig:amr_example_syntax} also shows the mechanics of framesets, where each frameset has a pre-defined set of slots. 
One example is the frameset "describe-01" has three pre-defined slots - :arg0 is the describer, :arg1 is the thing described, and :arg2 is what it is being described as \citep{banarescu-etal-2013-abstract}.

\begin{figure}[H]
    \centering
    \begin{minipage}{0.48\textwidth}
        \begin{lstlisting}{python}
# Sentence 1: The man  
# described the mission as a disaster.
# Sentence 2: The man's  
# description of the mission: disaster.
# Sentence 3: As the man described it, 
# the mission was a disaster.
(d / describe-01
    :arg0 (m / man)
    :arg1 (m2 / mission)
    :arg2 (d / disaster))
        \end{lstlisting}
        \caption{Example of Different English Sentences represented by the same AMR Graph \citep{banarescu-etal-2013-abstract}}
        \label{fig:amr_example_syntax}
    \end{minipage}
    \hfill
    \begin{minipage}{0.48\textwidth}
        \begin{lstlisting}{python}
# The boy from the college sang.
(s / sing-01 
 :arg0 (b / boy 
      :source (c / college)))

# The college boy who sang.
(b / boy 
 :arg0-of (s / sing-01) 
 :source (c / college))
        \end{lstlisting}
        \caption{Example of different AMR Graphs for similar sentences \citep{banarescu-etal-2013-abstract}}
        \label{fig:amr_example_syntax_2}
    \end{minipage}
\end{figure}

Secondly, seemingly similar sentences can be represented by different AMR Graphs because they contain different focuses in the sentence, as shown in Figure \ref{fig:amr_example_syntax_2}. AMR identifies the focus of the sentence by placing it at the top level root node \citep{banarescu-etal-2013-abstract}. Inverse relations, such as :arg0-of, enable us to change the focus of the rooted structure representations as needed \citep{inverserelations}. 

Lastly, a key feature of AMR is reentrancy \citep{szubert-etal-2020-role}, which allows concepts to participate in multiple relations simultaneously. As illustrated in Figure \ref{fig:amr_example}, a single concept like "boy" can serve as an argument to multiple predicates, following PropBank's systematic approach to semantic role labeling.

The AMR formalism has become increasingly popular in the Natural Language Processing (NLP) research community \citep{zabokrtsky-etal-2020-sentence, Bos2016SquibEP}. 
AMR has been successfully applied across numerous applications, including biomedical information extraction \citep{zhang-etal-2021-fine}, legal research \citep{vijay-hershcovich-2024-abstract}, machine translation \citep{song-etal-2019-semantic}, text summarisation \citep{dohare2017textsummarizationusingabstract, liu-etal-2015-toward, Liao2018AbstractMR}, sentence compression \citep{takase-etal-2016-neural}, 
text generation \citep{song-etal-2016-amr, song-etal-2018-graph, damonte-etal-2017-incremental, wang-etal-2020-amr, mager-etal-2020-gpt, zhao-etal-2020-bridging, fan-gardent-2020-multilingual, wang-etal-2021-better, bai2020onlinebackparsingamrtotextgeneration, jin-gildea-2020-generalized}, 
human-robot interaction and dialogue system natural language understanding \citep{bonial-etal-2020-dialogue, bonn-etal-2020-spatial} and event extraction \citep{li-etal-2020-cross, huang-etal-2016-liberal}.

\subsubsection{Overview of AMR Parsing Approaches}

Over the years, researchers have explored several distinct paradigms for AMR parsing, each reflecting evolving perspectives on how best to map natural language to structured meaning. Three main approaches have become established in the field, while a fourth, driven by recent advances in LLMs, is rapidly emerging as a promising new direction:

\paragraph{Graph-based Approaches} \citep{flanigan-etal-2014-discriminative, werling2015robustsubgraphgenerationimproves, lyu-titov-2018-amr, flanigan-etal-2016-cmu} frame parsing as a structured prediction task that directly optimises the entire graph structure. These methods typically decompose parsing into concept identification followed by relation prediction, employing global optimisation techniques to find maximum spanning connected subgraphs. However, their dependence on complex algorithms makes them difficult to scale, sensitive to error propagation, and less adaptable to diverse linguistic phenomena.

\paragraph{Transition-based Approaches} \citep{zhou-etal-2016-amr, damonte-etal-2017-incremental, wang-etal-2015-transition, ballesteros-al-onaizan-2017-amr, aaaiAMRParsing, naseem-etal-2019-rewarding} incrementally construct AMR graphs through a sequence of state transitions and actions. They maintain an explicit parser state (stack, buffer, and partial graph) and apply defined operations to transform this state until completion. However, while computationally efficient, their reliance on local, stepwise decisions can lead to compounding errors and makes it challenging to capture global context or long-range dependencies within the graph.

\paragraph{Sequence-to-Sequence Approaches} \citep{konstas-etal-2017-neural, bevilacqua2021one, zhang-etal-2019-amr} treat AMR parsing from natural language to a linearised representation of the AMR graph as a translation task. Unlike transition-based methods, they don't explicitly model intermediate states or transitions, instead using encoder-decoder architectures to map directly from input text to output AMR. However, the necessity to linearise inherently non-linear graph structures can result in information loss, difficulties in enforcing structural constraints, and challenges in accurately representing reentrancies and graph isomorphisms.

\paragraph{Decoder-Only Approaches} Therefore, the emergence of decoder-only LLMs (Phi, Gemma, LLaMA, DeepSeek-R1-LLaMA-Distilled) presents a promising new direction for AMR parsing. Unlike encoder-decoder architectures that compress information through bottlenecks \citep{vaswani2023attentionneed}, decoder-only models offer advantages for graph construction with their stacked layers and masked self-attention \citep{suresh2024smallerfasterdecoderonlytransformers}. 
These models maintain rich contextual representations throughout sequences \citep{hoffmann2022trainingcomputeoptimallargelanguage} and their autoregressive nature \citep{katharopoulos2020transformersrnnsfastautoregressive} suits the incremental construction of AMR graphs. When enhanced with GQA \citep{ainslie2023gqatraininggeneralizedmultiquery} for efficiency, Chain-of-Thought reasoning \citep{wei2023chainofthoughtpromptingelicitsreasoning} for explicit graph construction steps, and Parameter-Efficient LoRA fine-tuning \citep{hu2021lora} for resource-conscious adaptation, these models effectively overcome limitations in handling complex semantic structures and long-range dependencies.

\section{Methods}
This section provides a step-by-step overview of each part of the evaluation pipeline. 
Firstly, details of the training process are provided. This includes the dataset used, model choices, data preparation process, training parameters and training results.

Afterwards, details of the evaluation pipeline are provided. This includes the semantic and structural evaluation metrics used, generation parameters and output processing template.

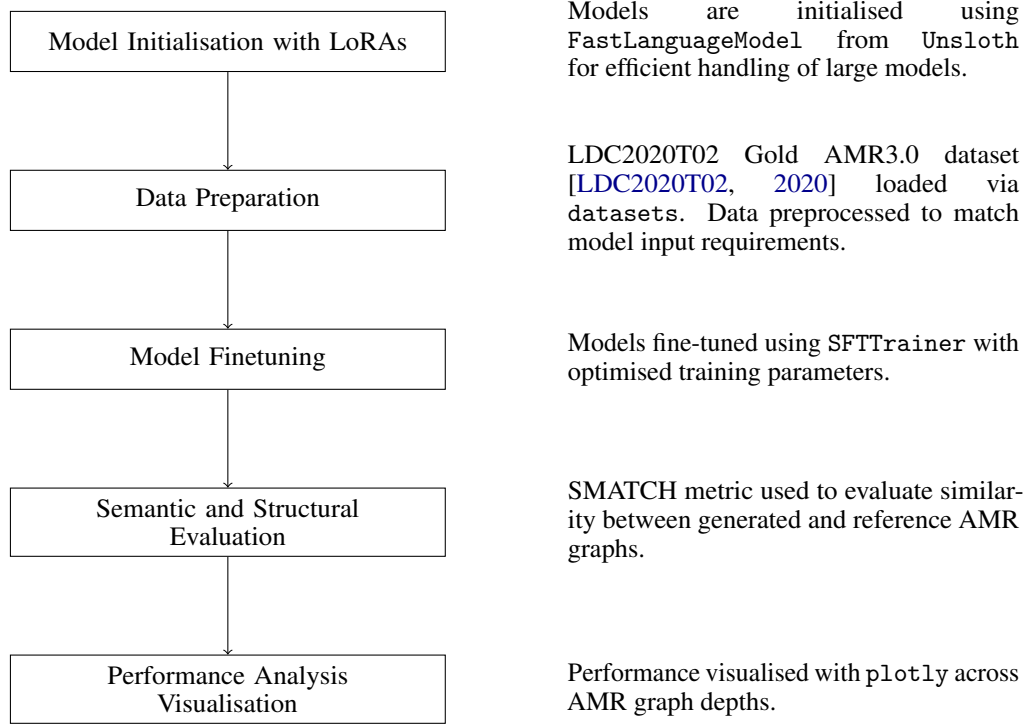
\begin{figure}[H]
\centering
\begin{tikzpicture}[
    node distance=1.3cm,
    box/.style={
        rectangle, 
        draw, 
        text width=5.5cm,
        align=center,
        minimum height=0.8cm,
        inner sep=4pt
    },
    desc/.style={
        text width=6cm,
        align=justify
    }
]
    \node[box] (init) {Model Initialisation with LoRAs};
    \node[box, below=of init] (data) {Data Preparation};
    \node[box, below=of data] (train) {Model Finetuning};
    \node[box, below=of train] (eval) {Semantic and Structural\\Evaluation};
    \node[box, below=of eval] (viz) {Performance Analysis\\Visualisation};
    
    \draw[->] (init.south) -- (data.north);
    \draw[->] (data.south) -- (train.north);
    \draw[->] (train.south) -- (eval.north);
    \draw[->] (eval.south) -- (viz.north);
    
    \node[desc, right=1.5cm of init] {Models are initialised using \texttt{FastLanguageModel} from \texttt{Unsloth} for efficient handling of large models.};
    
    \node[desc, right=1.5cm of data] {LDC2020T02 Gold AMR3.0 dataset \citep{LDC2020T02} loaded via \texttt{datasets}. Data preprocessed to match model input requirements.};
    
    \node[desc, right=1.5cm of train] {Models fine-tuned using \texttt{SFTTrainer} with optimised training parameters.};
    
    \node[desc, right=1.5cm of eval] {SMATCH metric used to evaluate similarity between generated and reference AMR graphs.};
    
    \node[desc, right=1.5cm of viz] {Performance visualised with \texttt{plotly} across AMR graph depths.};
\end{tikzpicture}
\caption{Evaluation Pipeline Components and Descriptions}
\label{fig:evaluation_pipeline}
\end{figure}

Figure \ref{fig:evaluation_pipeline} shows the training and evaluation flow that is applied to each model. 

\subsection{Dataset}
The AMR Annotation Release 3.0 \citep{LDC2020T02} serves as the primary dataset for this research. Developed by the Linguistic Data Consortium in collaboration with \href{https://www.rws.com/language-weaver/}{SDL/Language Weaver}, the \href{https://www.colorado.edu/lab/clear/}{University of Colorado}, and \href{http://www.isi.edu/home}{USC's Information Sciences Institute} \citep{Knight2020-uu}, this semantic treebank comprises 59,255 English sentences from diverse sources including web text and discussion forums \citep{Knight2020-uu}. 

This third-generation release builds upon previous versions with improvements such as expanded annotations and improved PropBank-style frames, making it a comprehensive resource for semantic parsing research.
Each sentence is paired with a tree-structured graph utilising "PropBank frames, non-core semantic roles, coreference, named entity annotation, modality, negation, and quantification" \citep{Knight2020-uu}. 

Table \ref{tab:dataset-subsets} presents train (93.9\%)/eval (2.9\%)/test (3.2\%) split and the detailed distribution of various data subsets across training, development, and test partitions. Figure \ref{fig:amr-example} shows an example of an AMR Graph-English Sentence Pair from the Dataset.

\begin{table}[!ht]
    \centering
    \begin{tabular}{>{\ttfamily}l<{} >{\ttfamily}r<{} >{\ttfamily}r<{} >{\ttfamily}r<{} >{\ttfamily}r<{}}
        \toprule
        \textbf{Dataset} & \textbf{Training} & \textbf{Dev} & \textbf{Test} & \textbf{Totals} \\
        \midrule
        BOLT DF MT & 1061 & 133 & 133 & 1327 \\
        Broadcast conversation & 214 & 0 & 0 & 214 \\
        Weblog and WSJ & 0 & 100 & 100 & 200 \\
        BOLT DF English & 7379 & 210 & 229 & 7818 \\
        DEFT DF English & 32915 & 0 & 0 & 32915 \\
        Aesop fables & 49 & 0 & 0 & 49 \\
        Guidelines AMRs & 970 & 0 & 0 & 970 \\
        LORELEI & 4441 & 354 & 527 & 5322 \\
        2009 Open MT & 204 & 0 & 0 & 204 \\
        Proxy reports & 6603 & 826 & 823 & 8252 \\
        Weblog & 866 & 0 & 0 & 866 \\
        Wikipedia & 192 & 0 & 0 & 192 \\
        Xinhua MT & 741 & 99 & 86 & 926 \\
        \midrule
        \textbf{Totals} & 55635 & 1722 & 1898 & 59255 \\
        \bottomrule
    \end{tabular}
    \caption{Dataset Subset Distribution}
    \label{tab:dataset-subsets}
\end{table}

\begin{figure}[!ht]
    \begin{lstlisting}[breaklines]{text}
    # ::id sdl_0002.2 ::date 2013-07-04T02:23:45 ::annotator SDL-AMR-09 ::preferred
    # ::snt This will ultimately accelerate the speed of desertification in sub - Saharan African countries and other areas of the world .
    # ::save-date Wed Feb 10, 2016 ::file sdl_0002_2.txt
    (a / accelerate-01
            :ARG0 (t / this)
            :ARG1 (s / speed-01
                :ARG0 t
                :ARG1 (d / desertification
                        :location (a2 / and
                            :op1 (c / country
                                    :location (w2 / world-region :wiki "Sub-Saharan_Africa" :name (n / name :op1 "Sub-Saharan" :op2 "Africa")))
                            :op2 (a3 / area
                                    :part-of (w / world)
                                    :mod (o / other)))))
            :time (u / ultimate))
\end{lstlisting}
\caption{Example of an AMR Graph-English Sentence Pair from the Dataset \citep{LDC2020T02}}
\label{fig:amr-example}
\end{figure}  

\subsection{Model Choices}
Four State-of-the-Art (SOTA) LLMs, Phi-3.5 \citep{abdin2024phi3technicalreporthighly}, Gemma-2 \citep{gemmateam2024gemma2improvingopen}, LLaMA-3.2 \citep{githubLLaMA3MODEL_CARDmdMain}, and DeepSeek-R1-LLaMA-Distilled \citep{deepseekai2025deepseekr1incentivizingreasoningcapability}, are finetuned and inferenced on the LDC2020T02 Gold AMR3.0 test set \citep{LDC2020T02}.

The selection of these compact models enables fine-tuning on consumer-grade hardware, specifically RTX 3060Ti GPUs, thereby obviating the need for high-end computational resources such as the A100. Each model represents a unique architectural approach relevant to the AMR parsing task.

LLaMA-3.2 (3B) incorporates Grouped-Query Attention (GQA), a mechanism designed to balance computational efficiency with strong performance. This architecture is particularly well-suited for processing complex AMR graphs while preserving semantic coherence. 
DeepSeek-R1-LLaMA-Distilled (8B) is distinguished by its integration of Chain of Thought reasoning, which facilitates step-by-step inference and self-correction during semantic analysis capabilities that are critical for accurately capturing complex semantic roles within AMR structures. 
Phi-3.5 (3.8B) provides robust multilingual support across 23 languages and features a 128K token context window, making it a promising candidate for future research in cross-lingual AMR parsing despite its relatively compact size. 
Gemma-2 (2B) leverages knowledge distillation and interleaved attention mechanisms to efficiently balance the modeling of local semantic details with the maintenance of global document structure, both of which are essential for comprehensive AMR graph construction.

By evaluating these four models, this research aims to provide a comparative analysis of their respective strengths and limitations in the context of AMR parsing, with a focus on both semantic and structural aspects.

\subsection{Model Initialisation}
Figure \ref{fig:lora_config} presents the details of our implementation. 
In this work, LoRA is configured with a set of parameters specifically chosen to balance parameter efficiency and model performance for the AMR parsing task \citep{Zhu2025-ha}. 
The rank parameter is set to $r=8$, representing a moderate value that provides sufficient model capacity for learning complex AMR structures while maintaining computational efficiency. 
The scaling factor, $\texttt{lora\_alpha}=32$, is selected to ensure that LoRA updates are strong enough to facilitate effective learning, yet not so large as to introduce instability during training. 

The configuration targets the attention components of the model, specifically the Query, Key, Value, and Output projections (\texttt{target\_modules} = [``q\_proj'', ``k\_proj'', ``v\_proj'', ``o\_proj'']), as these modules are critical for capturing the relationships between words and concepts inherent in AMR parsing. 
To mitigate overfitting while preserving the majority of learned features, a small dropout value of $\texttt{lora\_dropout}=0.05$ is applied. 

Collectively, these configuration choices are designed to optimise the model's ability to represent and generalise over the complex semantic structures present in the AMR dataset.

\begin{figure}[h!]
\centering
\begin{lstlisting}[breaklines, xleftmargin=5em, xrightmargin=2em]{python}
config = LoraConfig(
    r=8,                    # Rank dimension
    lora_alpha=32,          # Alpha parameter for LoRA scaling
    target_modules=[
        "q_proj",           # Query projection
        "k_proj",           # Key projection
        "v_proj",           # Value projection
        "o_proj"            # Output projection
    ],
    lora_dropout=0.05,      # Dropout probability
    bias="none",            # No bias parameters
    task_type="CAUSAL_LM"   # Task type for causal language modeling
)
\end{lstlisting}
\caption{LoRA Configuration Example \citep{hu2021lora}}
\label{fig:lora_config}
\end{figure}

\subsubsection{Model Loading Configuration}
The model loading configuration in this study is designed to maximise computational efficiency while preserving model performance. 
As illustrated in Figure \ref{fig:model_loading}, models are loaded using 4-bit quantisation (\texttt{load\_in\_4bit}), which substantially reduces the memory footprint by up to 75\% compared to standard 16-bit or 32-bit formats without significant loss in accuracy \citep{4bit}. 
This approach enables the fine-tuning of larger models on consumer-grade hardware. 

The numerical precision for model weights is set to bfloat16 (\texttt{bfloat16}), a format that offers an effective compromise between memory efficiency and numerical stability compared to full precision (float32) and standard half precision (float16). Furthermore, the maximum sequence length is set to 2048 tokens (\texttt{max\_seq\_length}), which is sufficient to accommodate the complexity of AMR graphs, particularly those with deep or highly nested structures. 

Collectively, these configuration choices facilitate the efficient and robust training of large language models for the AMR parsing task.

\begin{figure}[h!]
    \centering
    \begin{lstlisting}[breaklines, xleftmargin=5em, xrightmargin=2em]{python}
    model, tokenizer = FastLanguageModel.from_pretrained(
        model_name = "unsloth/LLaMA-3.2-3B",
        max_seq_length = 2048,
        dtype = bfloat16,
        load_in_4bit = True,
    )
    \end{lstlisting}
    \caption{Model Loading Configuration \citep{unsloth2024}}
    \label{fig:model_loading}
    \end{figure}
\subsection{Data Preparation}
\label{sec:data_prep}

The AMR 3.0 dataset is loaded using the \texttt{load\_dataset()} function from the Hugging Face \texttt{datasets} library \citep{huggingfaceDatasets}. 
This dataset contains pre-parsed AMR graphs paired with their corresponding natural language English sentences, providing a comprehensive resource for training AMR parsing models.

\begin{figure}[ht!]
\begin{lstlisting}[breaklines, xleftmargin=5em, xrightmargin=2em]{python}
def formatting_prompts_func(examples):
    system_prompt = "You are an AMR parser. Convert English sentences into Abstract Meaning Representation (AMR) graphs. Use proper AMR notation and formatting."

    texts = []
    for conversation in examples["conversations"]:
        messages = [
            {"role": "system", "content": system_prompt},
            conversation[0],  # user message
            conversation[1]   # assistant message
        ]
        text = tokenizer.apply_chat_template(
            messages,
            tokenize=False,
            add_generation_prompt=False
        )
        texts.append(text)
    return {"text": texts}

tokenizer = get_chat_template(
    tokenizer,
    chat_template="LLaMA-3.2",
)
dataset = dataset.map(formatting_prompts_func, batched=True)
\end{lstlisting}
\caption{Dataset Formatting Example \citep{huggingfaceDatasets}}
\label{fig:dataset_formatting}
\end{figure}

Figure \ref{fig:dataset_formatting} shows the dataset formatting process. This begins by defining a system prompt that specifies the role of the model as an AMR parser. This system prompt provides context for each conversation in the dataset.

The formatting function structures each dataset example into a three-part prompt: a system prompt defining the model as an AMR parser, the user's natural language English sentence, and the assistant's AMR graph response. Using LLaMA's chat template via \texttt{apply\_chat\_template()}, these conversations are properly formatted according to model-specific requirements. 

While Figure \ref{fig:dataset_formatting} demonstrates the LLaMA chat template, similar principles apply to other model architectures used in this project (Phi, Gemma, DeepSeek-R1-LLaMA-Distilled), each with their own specific chat templates. The process is efficiently applied to the entire dataset through batch processing with \texttt{dataset.map()}.

\subsection{Model Training}
Table \ref{tab:training-config-grouped} shows the parameters used in the model training process. The model training process uses the Hugging Face \texttt{SFTTrainer} class \citep{huggingfaceSupervisedFinetuning} with the following key parameters. 

\begin{table}[!ht]    
    \centering
    \begin{minipage}{0.65\textwidth}    
        \centering
        \begin{tabular}{>{\ttfamily}l<{} l >{\ttfamily}l<{} l}
            \toprule
            \multicolumn{4}{c}{\textbf{Training Configuration}} \\
            \midrule
            \multicolumn{2}{c}{\textit{Basic Parameters}} & \multicolumn{2}{c}{\textit{Optimisation Parameters}} \\
            \cmidrule(r){1-2} \cmidrule(l){3-4}
            batch\_size & 1 & learning\_rate & 2e-4 \\
            num\_epochs & 10 & optimiser & adamw\_4bit \\
            warmup\_steps & 10 & weight\_decay & 0.01 \\
            grad\_accum & 128 & lr\_scheduler & cosine \\
            \midrule
            \multicolumn{2}{c}{\textit{Logging Parameters}} & \multicolumn{2}{c}{\textit{Technical Parameters}} \\
            \cmidrule(r){1-2} \cmidrule(l){3-4}
            logging\_steps & 20 & fp16 & false \\
            save\_steps & 20 & bf16 & true \\
            eval\_steps & 20 & seed & 3407 \\
            save\_limit & 3 & report\_to & wandb \\
            \bottomrule
        \end{tabular}
        \caption{Training Configuration Parameters}
        \label{tab:training-config-grouped}
    \end{minipage}
\end{table}
\subsubsection{Training Results}
\label{sec:training_results}

\begin{figure}[H]
    \centering
    \includegraphics[width=0.8\textwidth]{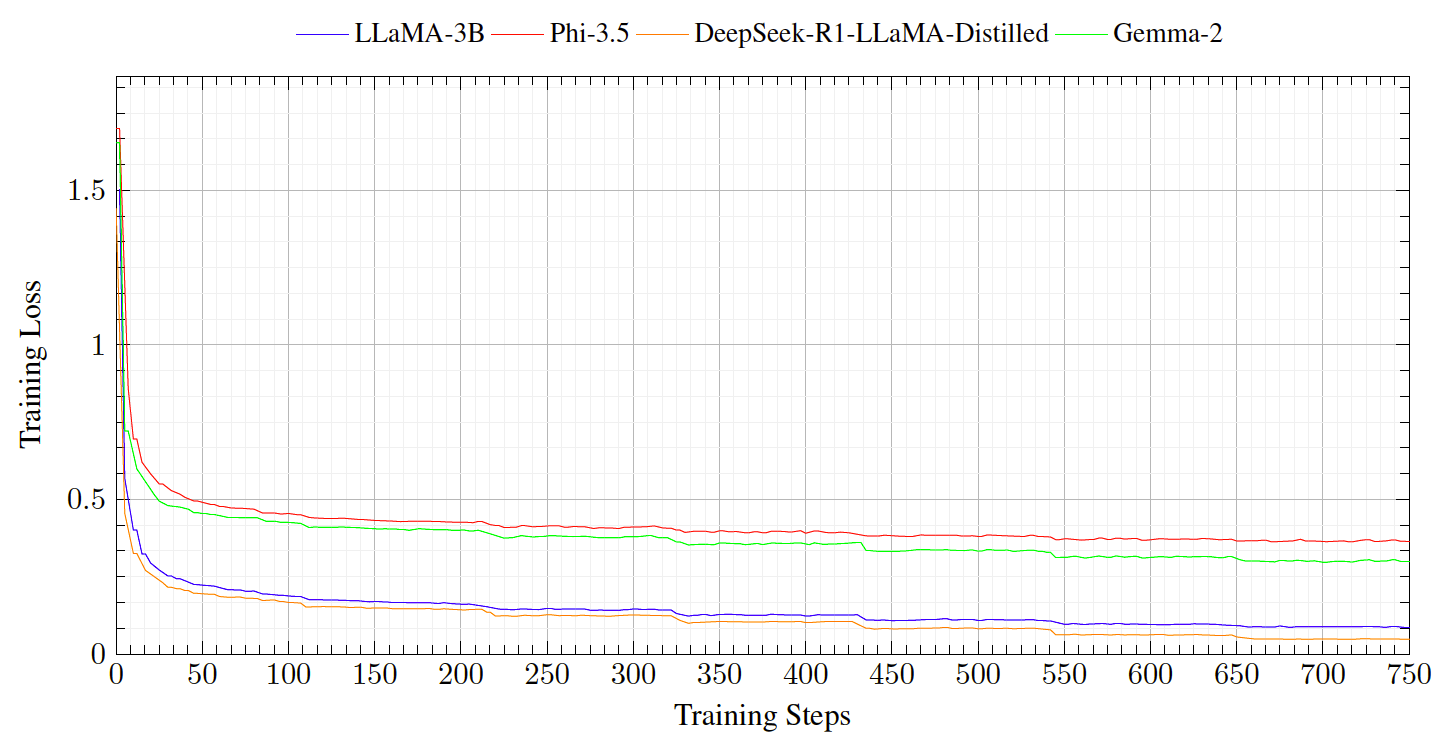}
    \caption{Training Loss}
    \label{fig:train_loss}
\end{figure}

\begin{figure}[H]
    \centering
    \includegraphics[width=0.8\textwidth]{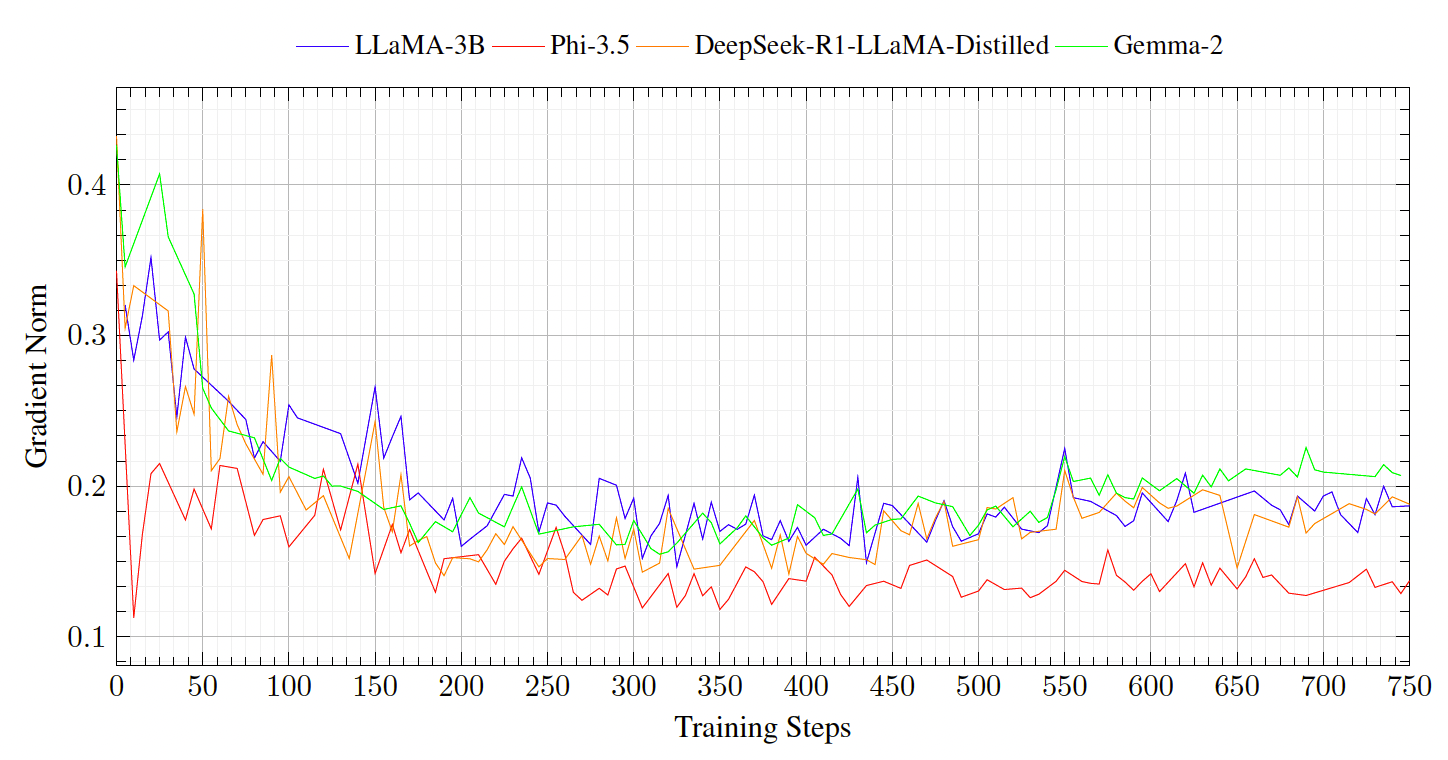}
    \caption{Training Gradient Norm}
    \label{fig:train_gradient_norm}
\end{figure}

\begin{figure}[H]
    \centering
    \includegraphics[width=0.8\textwidth]{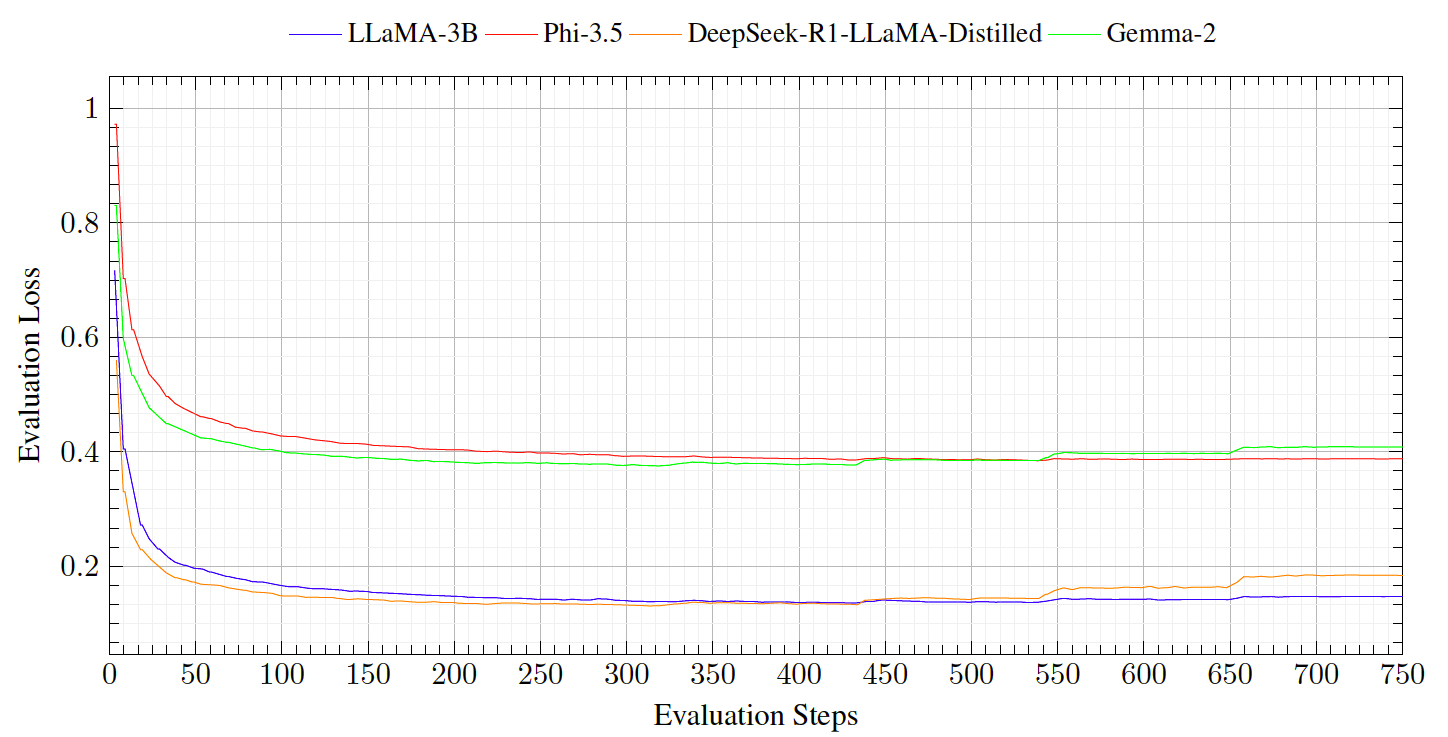}
    \caption{Evaluation Loss}
    \label{fig:eval_loss}
\end{figure}

Figures \ref{fig:train_loss}, \ref{fig:train_gradient_norm}, and \ref{fig:eval_loss} illustrate key training metrics captured on the integrated cloud-based WandB dashboard. 
Table \ref{tab:model-loss-comparison} presents the lowest training and evaluation loss across all models observed during training.
This real-time monitoring enables prompt detection of issues like divergence or overfitting, allowing for immediate hyperparameter adjustments when needed. Each data point corresponds to an evaluation step, providing detailed visibility into the model's learning progression.

\begin{table}[!ht]
    \centering
    \begin{tabular}{|l|c|c|c|c|}
        \hline
        \textbf{Metric} & \textbf{LLaMA-3.2-3B} & \textbf{Phi-3.5} & \textbf{DeepSeek-R1-LLaMA-Distilled} & \textbf{Gemma2-2B} \\
        \hline
        Training Loss & 0.0862 & 0.3638 & 0.0476 & 0.299 \\
        Evaluation Loss & 0.1355 & 0.3875 & 0.1302 & 0.3762 \\
        \hline
    \end{tabular}
    \caption{Lowest Training and Evaluation Loss Across Models}
    \label{tab:model-loss-comparison}
\end{table}
\subsection{SMATCH Evaluation Metric}
SMATCH (Semantic Match), by \citep{cai-knight-2013-smatch}, serves as the primary evaluation metric for assessing AMR parsing quality in this research. It is specifically designed to compare AMRs by treating them as semantic graphs and calculating the degree of overlap between the predicted and reference AMRs \citep{anchieta2019semaextendedsemanticevaluation}.

SMATCH operates by converting AMR expressions into triples (relation, source, target) that represent the graph structure \citep{opitz-etal-2020-amr}. SMATCH tackles the NP-hard problem of structural graph alignment \citep{springerComplexitySequence, opitz-2023-smatch} by finding the optimal mapping between variables in the predicted and reference graphs through a hill-climbing algorithm \citep{opitz-2023-smatch}.
Using these mappings, SMATCH computes Precision, Recall, and F1 scores based on the number of matching triples between the predicted and reference AMRs \citep{cai-knight-2013-smatch}.

We can see the advantages and limitations of SMATCH in Table \ref{tab:smatch_advantages_limitations}. In summary, SMATCH is a graph-centric metric that directly evaluates the semantic structure of AMRs, but it does not provide granular feedback for partially correct structures and may suffer from computational cost and variable agnosticism.

\subsubsection{Advantages and Limitations}
\begin{table}[H]
\begin{tabular}{p{0.45\textwidth}|p{0.45\textwidth}}
\hline
\textbf{Advantages} & \textbf{Limitations} \\
\hline
\textbf{Graph-centric}: Directly evaluates the semantic structure rather than surface-level similarities \citep{opitz-2023-smatch} & \textbf{Partial credit}: Does not provide granular feedback for partially correct structures \citep{groschwitz-etal-2023-amr} \\
\hline
\textbf{Industry standard}: Widely adopted in the AMR parsing community, enabling direct comparison with other research \citep{opitz-frank-2022-better} & \textbf{Computational cost}: The matching process can be computationally intensive for complex graphs with large number of variables \citep{naseem-etal-2022-docamr} \\
\hline
\textbf{Versatility}: Can be applied to different types of semantic formalisms, including AMR \citep{opitz-2023-smatch} & \textbf{Variable agnostic}: Focuses on maximising triple matches rather than considering the semantic content of the variables themselves \citep{song-gildea-2019-sembleu} \\
\hline
\end{tabular}
\caption{Advantages and Limitations of the SMATCH Evaluation Metric}
\label{tab:smatch_advantages_limitations}
\end{table}

\subsection{Inference Pipeline}
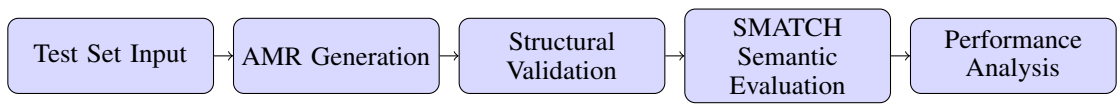
\begin{figure}[h]
    \centering
    \begin{tikzpicture}[
        node distance=3cm,
        block/.style={rectangle, draw, fill=blue!15, text width=2.5cm, text centered, minimum height=1cm, rounded corners}
    ]
        \node[block] (input) {Test Set Input};
        \node[block, right of=input] (generation) {AMR Generation};
        \node[block, right of=generation] (validation) {Structural Validation};
        \node[block, right of=validation] (evaluation) {SMATCH Semantic Evaluation};
        \node[block, right of=evaluation] (analysis) {Performance Analysis};
        
        \draw[->] (input) -- (generation);
        \draw[->] (generation) -- (validation);
        \draw[->] (validation) -- (evaluation);
        \draw[->] (evaluation) -- (analysis);
    \end{tikzpicture}
    \caption{AMR Inference Pipeline Architecture}
    \label{fig:inference_pipeline}
    \end{figure}

Figure \ref{fig:inference_pipeline} shows the inference pipeline for AMR parsing. The pipeline consists of multiple stages, from initial AMR generation, to structural and semantic validation, to detailed performance metrics across different depths and datasets. This modular approach enables both individual example evaluation and large-scale comparative analysis across different test sets.

\subsubsection{Generation Parameters}
The AMR is autoregressively generated until the model reaches the designated end-of-text token, with max\_length set to 2048. The generation process is governed by several key hyperparameters. 

The temperature is set to 0.7, which modulates the randomness of the output; this moderate value strikes a balance between deterministic generation (temperature approaching 0.0) and more diverse, but potentially less focused, outputs at higher temperatures. Top-p, or nucleus sampling, is configured at 1.0, thereby considering the entire probability distribution of possible tokens during generation. The repetition penalty is set to 1.0, ensuring that the model does not artificially penalise necessary word repetitions, thus supporting the production of natural and semantically coherent text.

\subsubsection{Template Structure}
The generation process utilises model-specific templates, corresponding to each model, for input and output processing. 

These inference prompt templates correspond to the prompt templates used in each model's fine-tuning phase, as seen in Figure \ref{fig:dataset_formatting} in Section \ref{sec:data_prep}. 
This ensures distributional consistency between the fine-tuning and inference phases, thus minimising representation drift.

The template structure for each model can be generalised as comprising three main components: a system message, a user message, and an assistant message. The system message establishes the context and constraints for the model's response generation, effectively priming the attention mechanism for the task. The user message contains the source text intended for AMR parsing, serving as the conditional input to the autoregressive decoder. Finally, the assistant message represents the target space in which the model generates the AMR graph through next-token prediction.

\subsubsection{Output Processing}
The generated responses contain AMR graph representations interspersed with model-specific template token delimiters. These delimiters vary across architectures: LLaMA-3.2 and DeepSee-R1-LLaMA-Distilled use \texttt{<|start\_header\_id|>} and \texttt{<|end\_header\_id|>} as shown in Figure \ref{fig:template_markers}, while Phi-3 employs \texttt{<|system|>, <|user|>, <|assistant|>} tokens, and Gemma uses \texttt{<bos>, <start\_of\_turn>, <end\_of\_turn>} tokens. The post-processing pipeline applies model-specific regex pattern matching to extract the canonical AMR representation from the tokenised output sequence.

\begin{figure}[!ht]
\begin{lstlisting}[breaklines]{python}
generated_text = self.tokenizer.decode(outputs[0], skip_special_tokens=False)
        
amr_start = generated_text.find("<|start_header_id|>assistant<|end_header_id|>")
if amr_start != -1:
    generated_amr = generated_text[amr_start:].split("<|eot_id|>")[0].strip()
    return generated_amr.replace("<|start_header_id|>assistant<|end_header_id|>", "").strip()
return generated_text
\end{lstlisting}
\caption{AMR Extraction between Template Markers}
\label{fig:template_markers}
\end{figure}

\section{Evaluation}
This section provides the inference results of our finetuned models. 
We first compare the performance of our finetuned models against SOTA models on the same Gold AMR 3.0 LDC2020T02 dataset.
We then conduct a depth analysis of the finetuned models individually during inference.

Afterwards, we run inference on a larger Silver MBSE dataset to further validate the performance of the finetuned models.
Lastly, we run inference on a subset of the Gold AMR3.0 LDC2020T02 dataset to investigate the performance of the finetuned models on different subsets within the dataset.

\subsection{Comparison with State-of-the-Art (SOTA) AMR Parsers}
\label{sec:comparison-with-state-of-the-art}
\begin{figure}[H]
    \centering
    \includegraphics[width=1.0\textwidth]{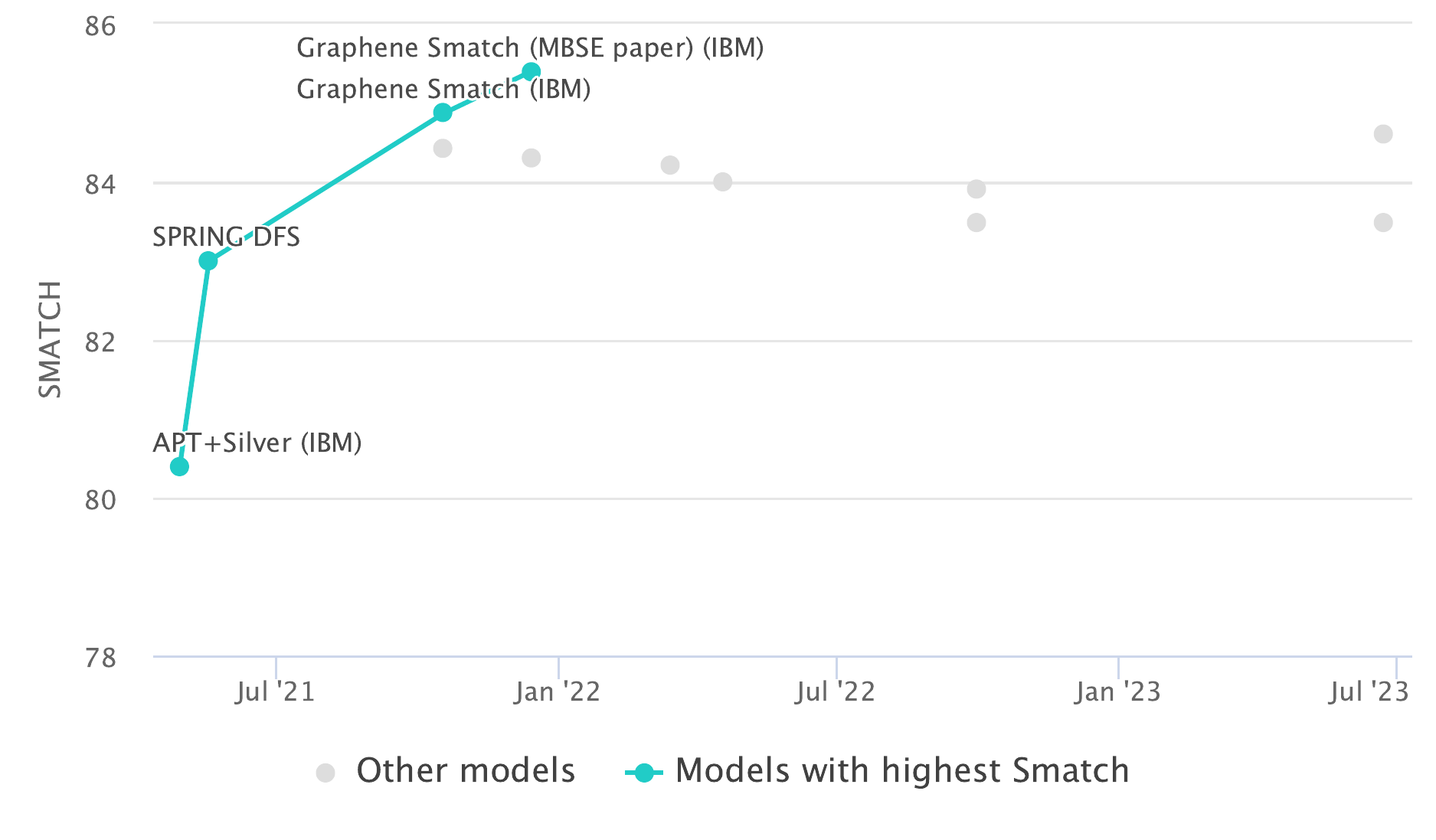}
    \caption{SOTA AMR Parsers Comparison \citep{paperswithcodePapersWith}}
    \label{fig:sota_comparison}
\end{figure}

We begin preliminary analysis by using the high-quality Gold AMR 3.0 LDC2020T02 dataset \citep{LDC2020T02}.
Figure \ref{fig:sota_comparison} shows the performance of SOTA models trained and tested on the same LDC2020T02 dataset \citep{LDC2020T02}, enabling fair comparison to our finetuned models.
Table \ref{tab:model-metrics} shows the performance of our finetuned models on the full 1722 samples in the LDC2020T02 dataset test split.

The F1, Precision, and Recall values are the means and confidence intervals of the performance metrics. 
The best scores for each metric are highlighted in bold.
The confidence intervals are calculated at the 95\% confidence level using the standard error of the mean across the depth levels, reflecting the variability in model performance across different graph complexities.

As seen in Figure \ref{fig:sota_comparison} and Table \ref{tab:model-metrics}, our results demonstrate remarkable competitiveness despite using a straightforward fine-tuning approach. 
Our best-performing model, LLaMA-3.2 (whose scores are highlighted in bold), achieves an F1 score of 0.804, which is comparable to APT + Silver (IBM) at 0.804 \citep{zhou2021amr}. 
While our models fall slightly short of the top SOTA models - Graphene Smatch (MBSE) at 0.854 \citep{lee-etal-2022-maximum} and Graphene Smatch (IBM) at 0.8487 \citep{hoang2021ensembling}, and Spring DFS at 0.830 \citep{bevilacqua2021one}), this performance gap is notably small considering that our models are general-purpose language models adapted through fine-tuning, rather than specialised architectures designed specifically for AMR parsing. 

The consistent performance across all three metrics (F1, Precision, and Recall) for each model also indicates robust and balanced parsing capabilities. 
This achievement is particularly noteworthy as it suggests that large language models can effectively tackle specialised tasks like AMR parsing with minimal architectural modifications.

\begin{table}[H]
    \centering
    \begin{tabular}{|l|c|c|c|c|}
        \hline
        \textbf{SMATCH Metric} & \textbf{LLaMA 3.2} & \textbf{DeepSeek-R1-LLaMA-Distilled} & \textbf{Gemma 2} & \textbf{Phi 3.5} \\
        \hline
        F1 & \textbf{0.804 ± 0.019} & 0.783 ± 0.020 & 0.793 ± 0.019 & 0.779 ± 0.020 \\
        \hline
        Precision & \textbf{0.812 ± 0.019} & 0.792 ± 0.019 & 0.818 ± 0.018 & 0.791 ± 0.020 \\
        \hline
        Recall & \textbf{0.801 ± 0.019} & 0.780 ± 0.020 & 0.775 ± 0.020 & 0.772 ± 0.020 \\
        \hline
    \end{tabular}
    \caption{Performance Metrics of Our Models}
    \label{tab:model-metrics}
\end{table}

\subsection{Depth Analysis}
We then continue analysis using the LDC2020T02 Gold AMR 3.0 dataset by analysing the performance of the four models across all depth levels.
The depth of an AMR graph refers to the maximum distance, in terms of number of edges, from the root node to any leaf node \citep{wang-etal-2020-amr}. This serves as a proxy for semantic complexity and hierarchical structure.

Because we want to have an equal number of samples per depth for fair comparison, we are limited to using 30 samples per depth level for depth 1-10 as there are fewer samples available for deeper depths. 

Table \ref{tab:model-analysis} presents a consolidated performance analysis of the four models across all depth levels.
The F1, Precision, and Recall values are the means and confidence intervals of the performance metrics across the 10 depth levels.
The Mean Error Count is the mean number of errors across the 10 depth levels.

The best scores for each metric are highlighted in bold.
The confidence intervals are calculated at the 95\% confidence level using the standard error of the mean across the depth levels, reflecting the variability in model performance across different graph complexities.

\begin{table}[!ht]
\resizebox{\textwidth}{!}{%
\begin{tabular}{|l|c|c|c|c|}
\hline
\multirow{2}{*}{\textbf{Model}} & \multicolumn{3}{c|}{\textbf{Semantic Validation}} & \textbf{Structural Validation} \\
\cline{2-5}
 & F1 Score & Precision & Recall & Mean Error Count \\
\hline
Gemma-2 & 0.79 ± 0.06 & 0.80 ± 0.05 & 0.78 ± 0.07 & 1.9 \\
\hline
DeepSeek-R1-LLaMA-Distilled & 0.83 ± 0.06 & 0.84 ± 0.05 & 0.83 ± 0.07 & 0.4 \\
\hline
LLaMA-3.2 & \textbf{0.87 ± 0.07} & \textbf{0.88 ± 0.06} & \textbf{0.87 ± 0.07} & 0.7 \\
\hline
Phi-3.5 & 0.79 ± 0.06 & 0.80 ± 0.05 & 0.79 ± 0.07 & \textbf{0.3} \\
\hline

\end{tabular}
}
\caption{Individual Model Performance Analysis}
\label{tab:model-analysis}
\end{table}

As seen in Table \ref{tab:model-analysis}, LLaMA-3.2 leads in semantic performance (F1: 0.87, Precision: 0.88, Recall: 0.87) but shows moderate structural errors (0.7). DeepSeek-R1-LLaMA-Distilled balances strong semantic results (F1: 0.83) with excellent structural validity (0.4 errors). Phi-3.5, despite its smaller size, achieves the best structural validity (0.3 errors). Gemma-2 performs weakest overall with the lowest semantic scores and highest structural error rates (1.9).

When comparing these results with our full test split evaluation (Section \ref{sec:comparison-with-state-of-the-art}, Table \ref{tab:model-metrics}), we observe that the depth analysis shows higher F1 scores (0.87 vs 0.804 for LLaMA-3.2, 0.83 vs 0.783 for DeepSeek-R1-LLaMA-Distilled, 0.79 vs 0.793 for Gemma-2, 0.79 vs 0.779 for Phi-3.5), likely due to the smaller, more focused sample size in the depth analysis.

\subsubsection{Summarised Analysis}
\label{sec:depth-analysis-summarised}

\begin{figure}[H]
    \centering
    \includegraphics[width=0.8\textwidth]{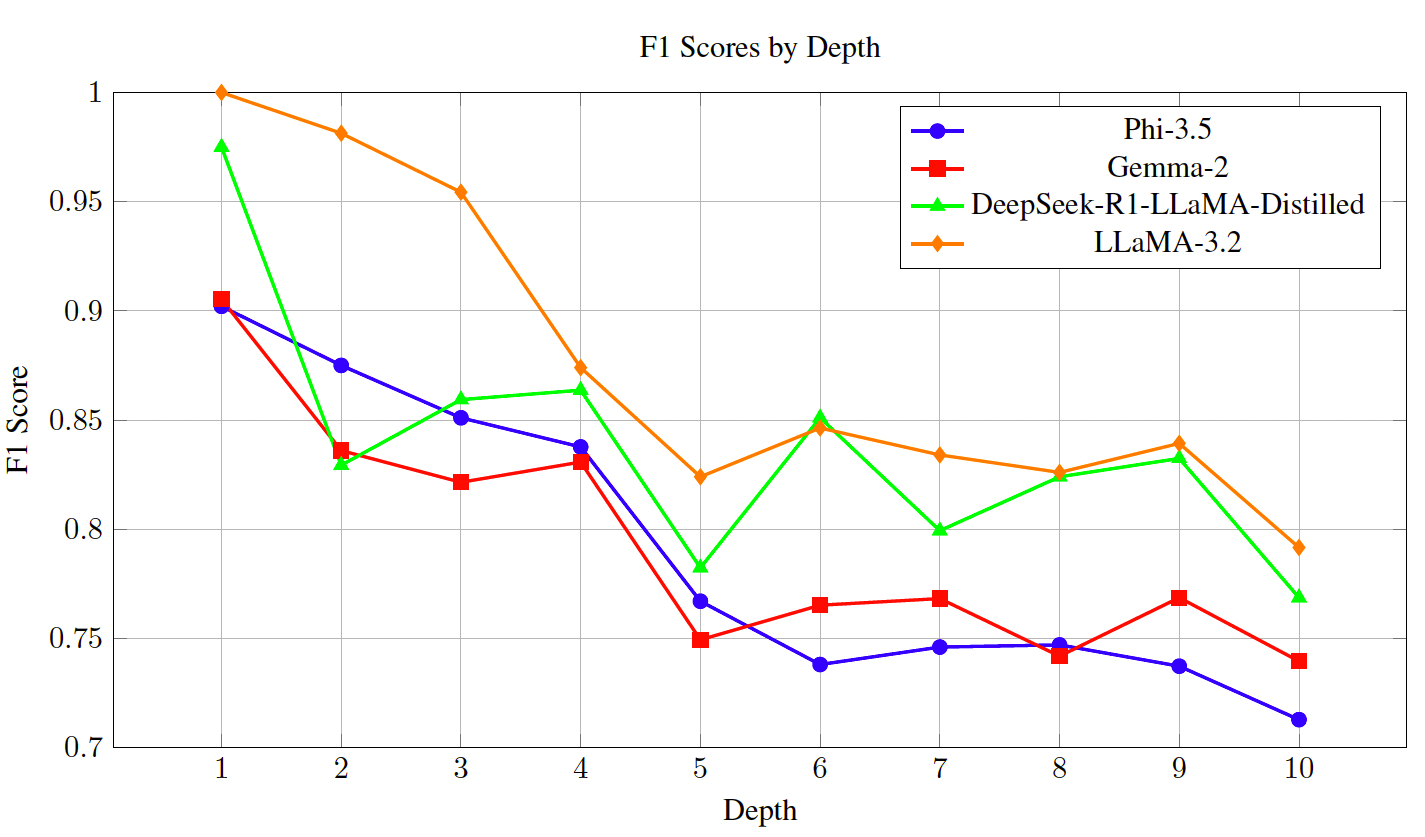}
    \caption{Consolidated F1 Scores [Gold AMR3.0 dataset]}
    \label{fig:consolidated_f1_indiv}
\end{figure}

\textbf{Semantic Validation}: LLaMA-3.2 leads in semantic performance (F1: 0.87, Precision: 0.88, Recall: 0.87) as shown in Table \ref{tab:model-analysis}. Figure \ref{fig:consolidated_f1_indiv} demonstrates LLaMA-3.2's consistent F1 range (0.79-1.0) across all depths, with peak performance at depth 9. This suggests that strong language modeling capabilities, rather than instruction tuning, are crucial for semantic understanding.

\begin{figure}[H]
    \centering
    \includegraphics[width=0.8\textwidth]{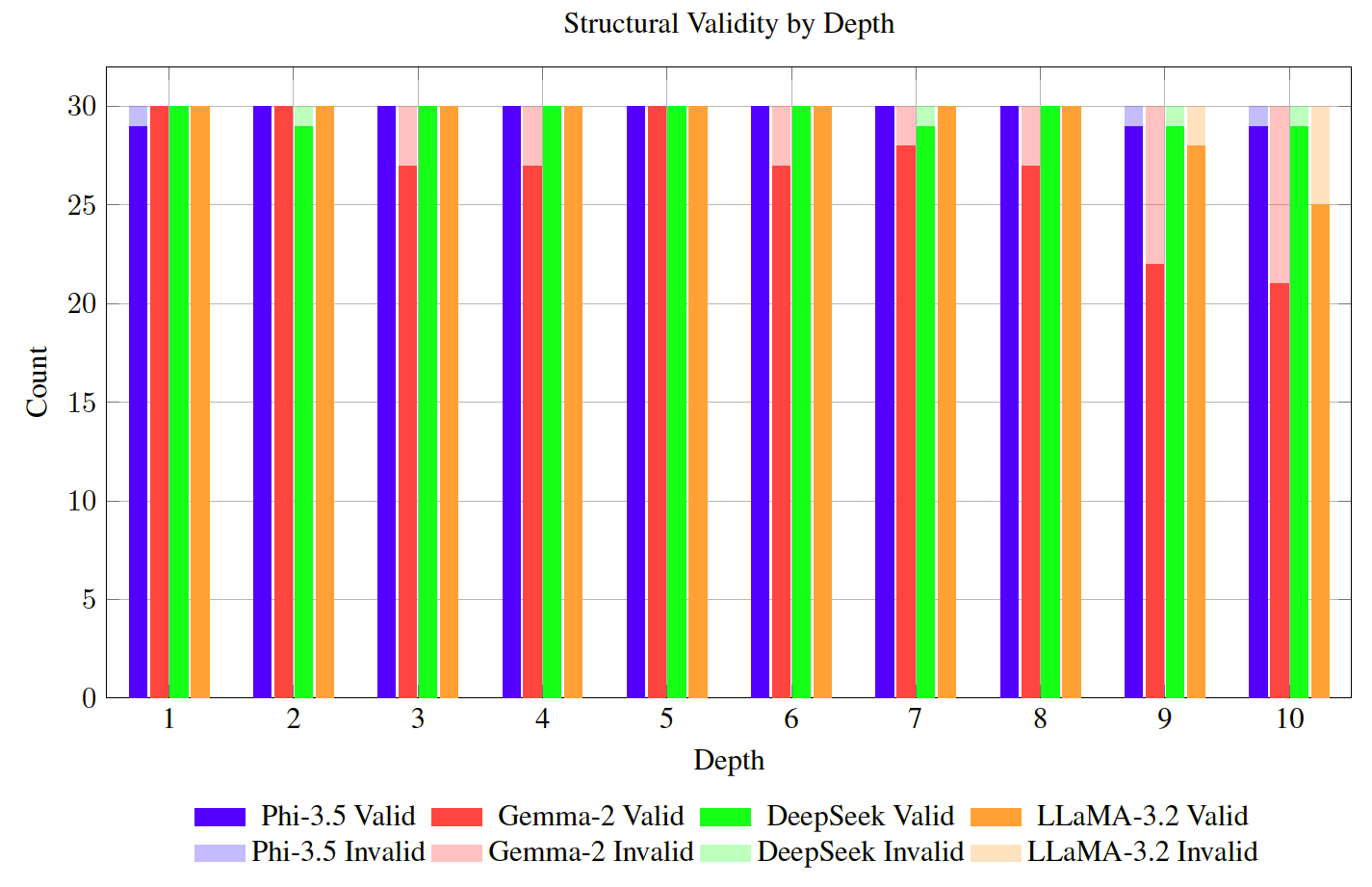}
    \caption{Consolidated Structural Validity [Gold AMR3.0 dataset]}
    \label{fig:consolidated_structural_indiv}
\end{figure}

\textbf{Structural Validation}: Phi-3.5 leads in structural performance with the lowest Mean Error Count (0.3) followed by DeepSeek-R1-LLaMA-Distilled (0.4) as shown in Table \ref{tab:model-analysis}. 
One notable exception, as seen in Figure \ref{fig:consolidated_structural_indiv}, is LLaMA-3.2, which maintained perfect structural validity through depths 1-8 but experienced rapid deterioration at depths 9-10. 
However, Phi-3.5 still maintains most consistent structural validity across all depths. 
This could be attributed to Phi-3.5's Instruction tuning pre-training method which enhances structural extraction capabilities, potentially at the expense of some semantic modeling capacity.

\begin{figure}[H]
    \centering
    \includegraphics[width=0.8\textwidth]{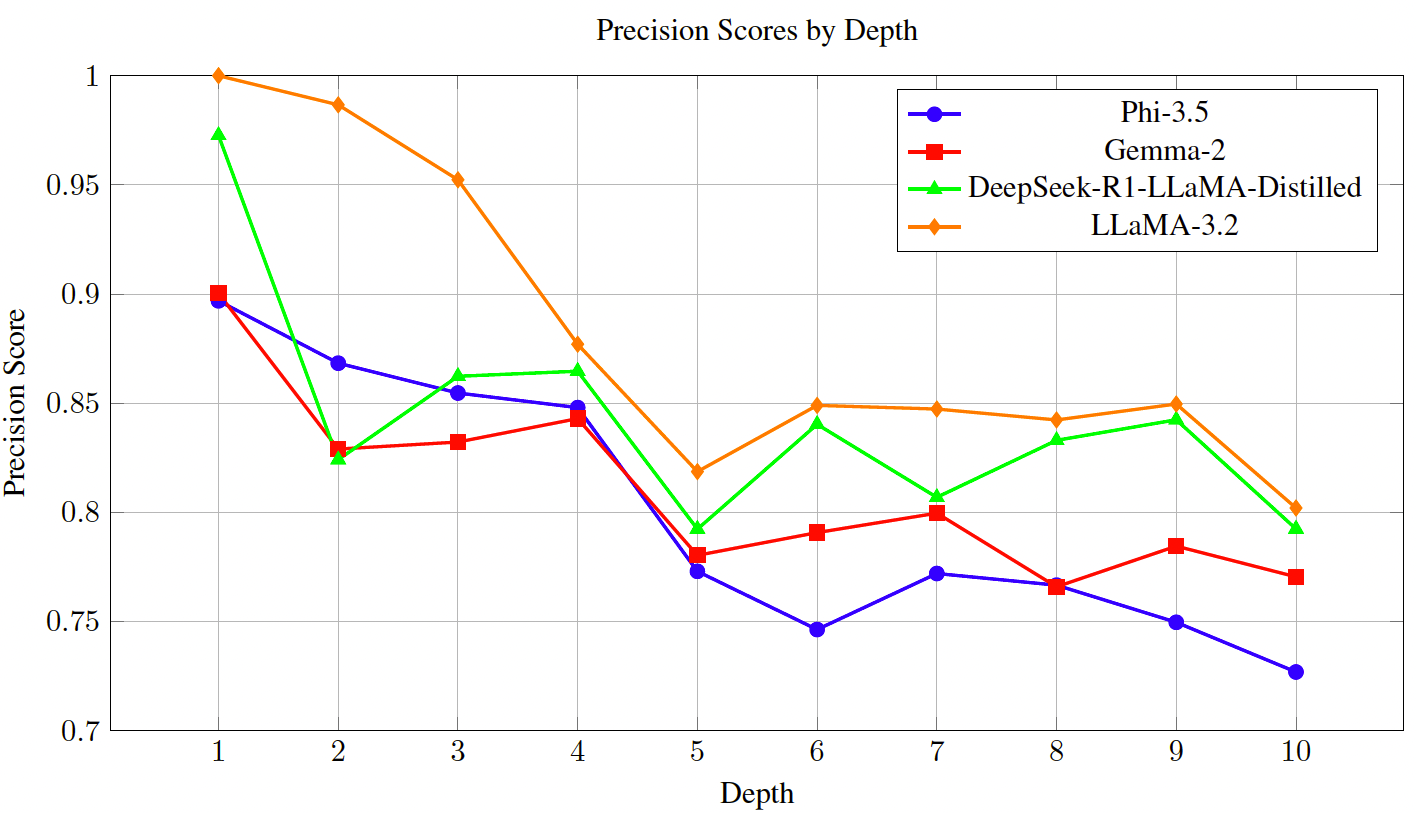}
    \caption{Consolidated Precision Scores [Gold AMR3.0 dataset]}
    \label{fig:consolidated_precision_indiv}
\end{figure}

\begin{figure}[H]
    \centering
    \includegraphics[width=0.8\textwidth]{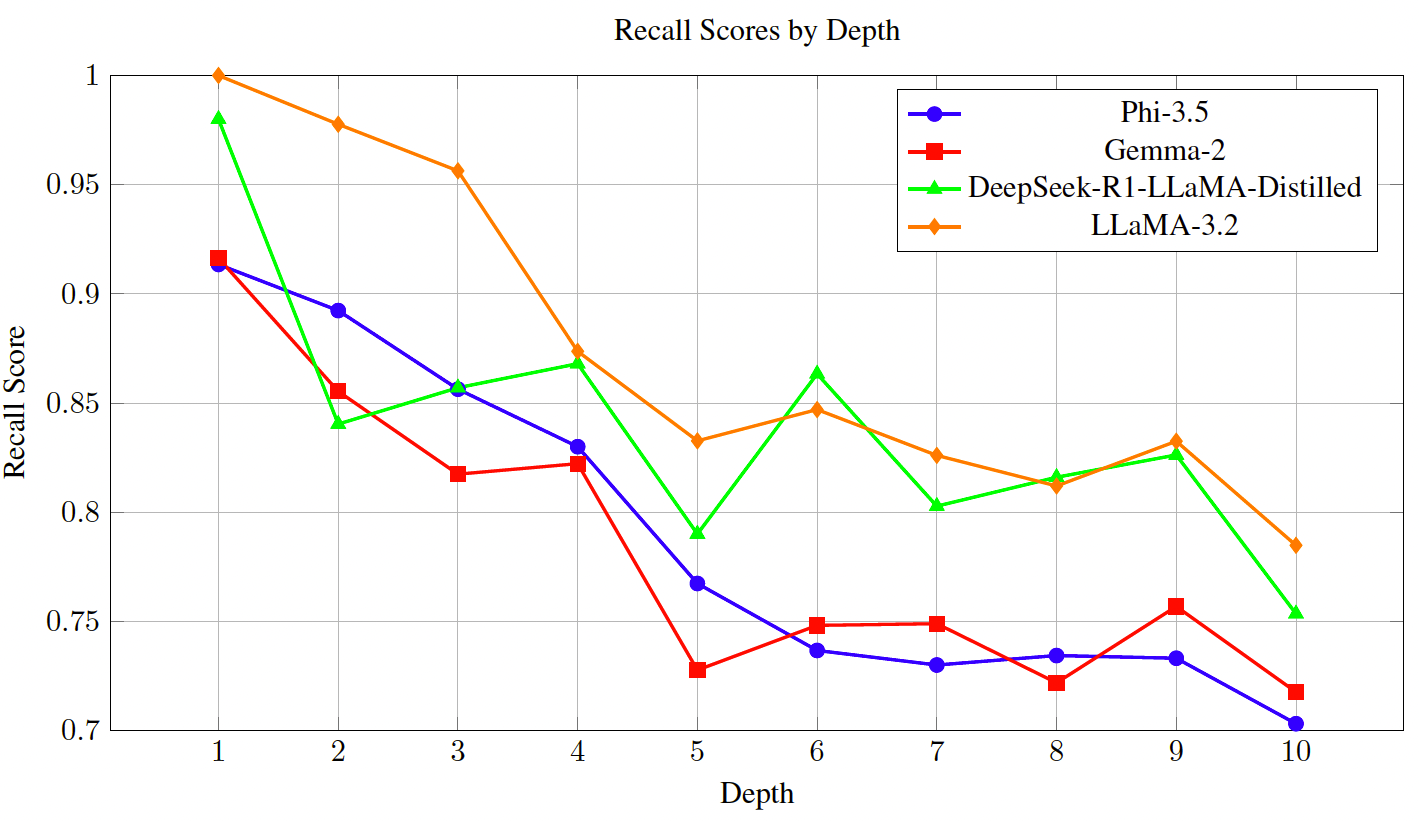}
    \caption{Consolidated Recall Scores [Gold AMR3.0 dataset]}
    \label{fig:consolidated_recall_indiv}
\end{figure}

\subsubsection{Structural Validation across Depth Analysis}
As shown in Figure \ref{fig:consolidated_structural_indiv}, our analysis of structural validation across AMR graph depths reveals three key observations. 

Firstly, LLaMA-3.2 maintains perfect structural validity through depths 1-8 but experiences significant deterioration at depths 9-10, revealing a threshold limitation in its pretrained language capabilities when handling the most complex hierarchical relationships.
    
Secondly, Phi-3.5, despite having a smaller parameter count (3.8B), achieves the most consistent structural validity across all depth levels. 
    
Lastly, Gemma-2 exhibits substantial structural errors at multiple depths (3, 4, 6, 7, 8) with error rates increasing dramatically at depths 9-10.

\subsubsection{Semantic Validation across Depth Analysis}
\label{sec:semantic-validation-depth-analysis}

As illustrated in Figures \ref{fig:consolidated_f1_indiv}, \ref{fig:consolidated_precision_indiv}, and \ref{fig:consolidated_recall_indiv}, our analysis of semantic validation across AMR graph depths reveals three key patterns. 

Firstly, across all models, Precision consistently exceeds Recall, indicating that current AMR parsing approaches are more effective at generating accurate graph fragments than capturing complete reference semantics.
    
Secondly, LLaMA-3.2 and DeepSeek-R1-LLaMA-Distilled, despite sharing architectural foundations, exhibit distinct semantic profiles: LLaMA-3.2 shows superior robustness with consistently higher F1 scores (average +0.09) and gradual degradation across depths, while DeepSeek-R1-LLaMA-Distilled excels at depth 1 (F1=0.975) but experiences pronounced fluctuations at intermediate depths—suggesting fundamental differences in their semantic extraction capabilities when handling complex hierarchical relationships.
    
Lastly, Phi-3.5 and Gemma-2 reveal how training approaches affect semantic performance: Phi-3.5's clear performance cliff after depth 4 (F1 drops from 0.838 to 0.767) suggests its instruction tuning approach prioritises structural validity at the expense of semantic performance beyond certain complexity thresholds. Meanwhile, Gemma-2's inconsistent results with fluctuating F1 scores across depths reinforce research findings on its historically known fine-tuning challenges \citep{springer2025overtrained}.

\subsection{Silver Data Analysis}
We evaluate our model using a silver dataset of 362,000 sentence-AMR pairs generated through Maximum Bayes Smatch Ensemble Distillation (MBSE) \citep{lee-etal-2022-maximum}. 

The \href{https://github.com/nj19257/AMR-silver-data-for-ensemble-distill}{Silver MBSE dataset} is particularly valuable as it was generated using multiple AMR3-structbart-Large instances, achieving high accuracy scores (85.9 on AMR2.0 and 84.3 on AMR3.0) \citep{lee-etal-2022-maximum}, and covers a broader range of linguistic phenomena than our gold test sets.

While not as authoritative as Gold data, this large-scale dataset enables several important avenues of analysis. First, it allows for scale validation by testing the consistency and reliability of model performance across a dataset that is substantially larger than standard AMR 3.0 test sets. Second, it facilitates the assessment of ensemble agreement, providing a means to evaluate the degree of alignment between model-generated AMRs and those produced by ensemble methods. Finally, the breadth of the dataset supports a more comprehensive analysis of error patterns, enabling the identification of systematic errors that may not be apparent in smaller test sets.

While our gold AMR 3.0 analysis used 30 samples per depth level, the silver dataset enables more robust statistical validation with 500 samples per depth level across depths 2-12 (based on data availability). 
By treating this Silver MBSE dataset as a gold standard for comparison, we significantly increase our confidence in depth-specific performance metrics and can better identify differing patterns in model behavior across different complexity levels.

Table \ref{tab:model-silver-analysis} presents the performance of the four models on the Silver MBSE dataset. 

The F1, Precision, and Recall values are the mean and confidence intervals of the performance metrics across the 10 depth levels.
The Mean Error Count is the mean number of errors across the 10 depth levels.

The best scores for each metric are highlighted in bold.
The confidence intervals are calculated at the 95\% confidence level using the standard error of the mean across the depth levels, reflecting the variability in model performance across different graph complexities.

\begin{table}[!ht]
\resizebox{\textwidth}{!}{%
\begin{tabular}{|l|c|c|c|c|}
\hline
\multirow{2}{*}{\textbf{Model}} & \multicolumn{3}{c|}{\textbf{Semantic Validation}} & \textbf{Structural Validation} \\
\cline{2-5}
 & F1 Score & Precision & Recall & Mean Error Count \\
\hline
Gemma-2 & 0.78 ± 0.02 & 0.78 ± 0.01 & 0.79 ± 0.03 & 32.6 \\
\hline
DeepSeek-R1-LLaMA-Distilled & 0.78 ± 0.05 & 0.77 ± 0.05 & 0.80 ± 0.05 & 7.3 \\
\hline
LLaMA-3.2 & \textbf{0.81 ± 0.06} & \textbf{0.80 ± 0.06} & \textbf{0.83 ± 0.06} & 20.2 \\
\hline
Phi-3.5 & 0.75 ± 0.03 & 0.74 ± 0.02 & 0.77 ± 0.03 & \textbf{6.3} \\
\hline
\end{tabular}
}
\caption{Individual Model Performance Analysis on Silver Dataset}
\label{tab:model-silver-analysis}
\end{table}

As seen in Table \ref{tab:model-silver-analysis}, LLaMA-3.2 leads in semantic performance (F1: 0.81, Precision: 0.80, Recall: 0.83) despite moderate structural errors (20.2). Conversely, Phi-3.5 achieves superior structural validity (6.3 errors) but the lowest semantic scores (F1: 0.75, Precision: 0.74, Recall: 0.77). DeepSeek-R1-LLaMA-Distilled balances competitive semantic metrics (F1: 0.78, Precision: 0.77, Recall: 0.80) with strong structural validity (7.3 errors), while Gemma-2 shows mid-range semantic performance (F1: 0.78, Precision: 0.78, Recall: 0.79) but poor structural validity (32.6 errors). These findings suggest a trade-off between semantic accuracy and structural coherence across models.

\subsubsection{Summarised Analysis}
\label{sec:silver-analysis-summarised}
\textbf{Semantic Validation}: LLaMA-3.2 leads semantically (F1: 0.81, Precision: 0.80, Recall: 0.83) as shown in Table \ref{tab:model-silver-analysis}, reinforcing our findings in Section \ref{sec:depth-analysis-summarised}. 
Interestingly, the Silver MBSE dataset shows Recall exceeding Precision across all models—contrary to patterns observed with Gold AMR3.0 data, perhaps due to the larger sample size (500 versus 30 samples per depth). 
Another interesting observation is that at the highest complexity levels (depths 11-12), all models converge to similar F1 scores (range of 0.03, from 0.72-0.75), compared to a wider range of 0.13 at depth 2. This suggests architectural and training differences become less relevant at extreme complexity levels, potentially indicating a fundamental ceiling in transformer-based approaches to complex semantic structures.

\begin{figure}[H]
    \centering
    \includegraphics[width=0.8\textwidth]{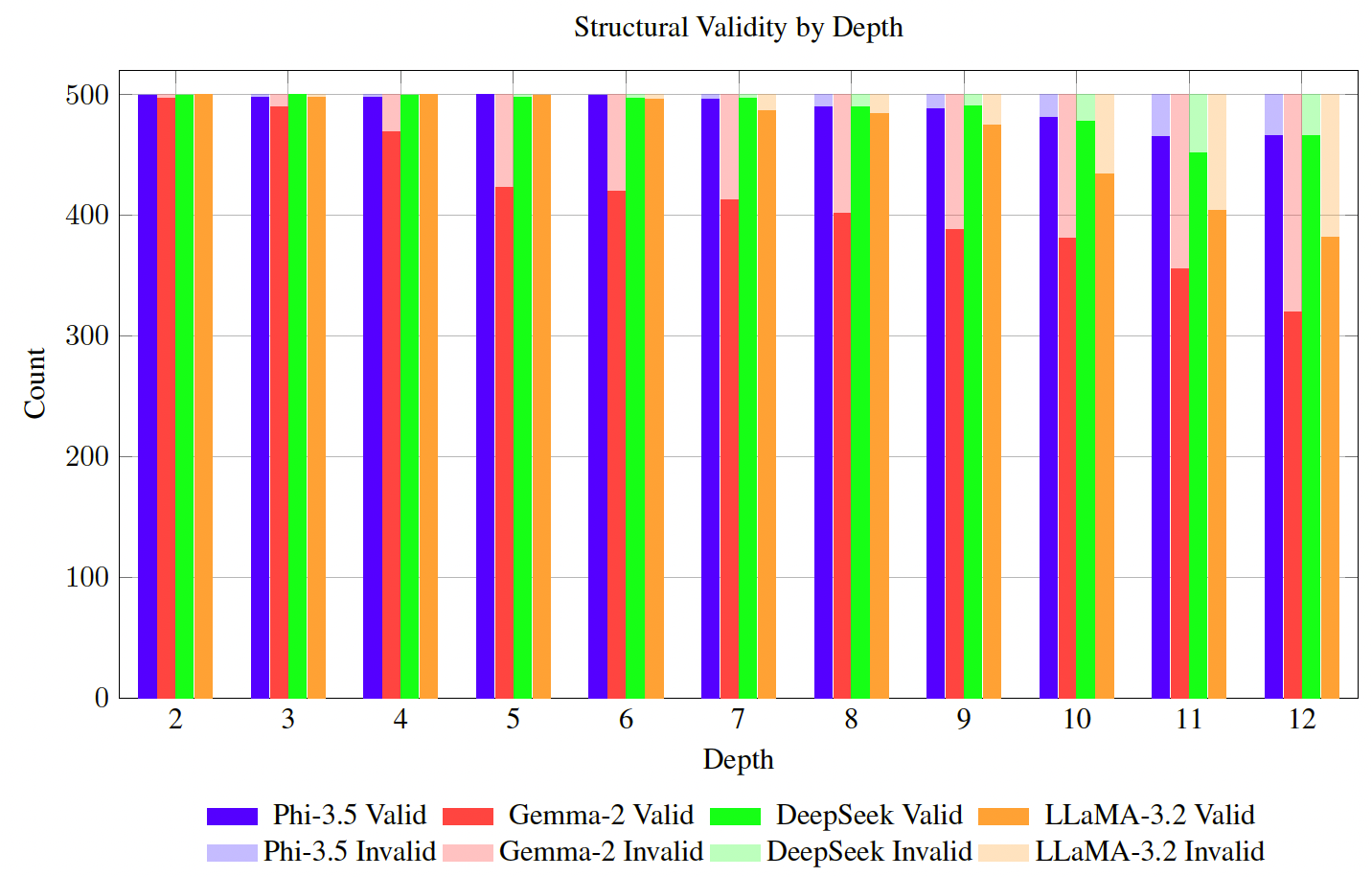}
    \caption{Consolidated Structural Validity [Silver MBSE AMR3.0 Dataset]}
    \label{fig:consolidated_structural_silver}
\end{figure}

\textbf{Structural Validation}: Phi-3.5 leads in structural performance with the lowest Mean Error Count (6.3) followed by DeepSeek-R1-LLaMA-Distilled (7.3) as shown in Table \ref{tab:model-silver-analysis}. Figure \ref{fig:consolidated_structural_silver} reveals Phi-3.5's exceptional structural robustness across all depths, maintaining validity rates above 93\% even at depth 12 (466/500 valid). 
This finding strongly reinforces our observation in Section \ref{sec:depth-analysis-summarised} that Phi-3.5 maintains the most consistent structural validity across all depths, further reinforcing the conclusion that its instruction-tuning approach is particularly effective for preserving structural coherence in complex graphs, even when tested on a much larger dataset.

\begin{figure}[H]
    \centering
    \includegraphics[width=0.8\textwidth]{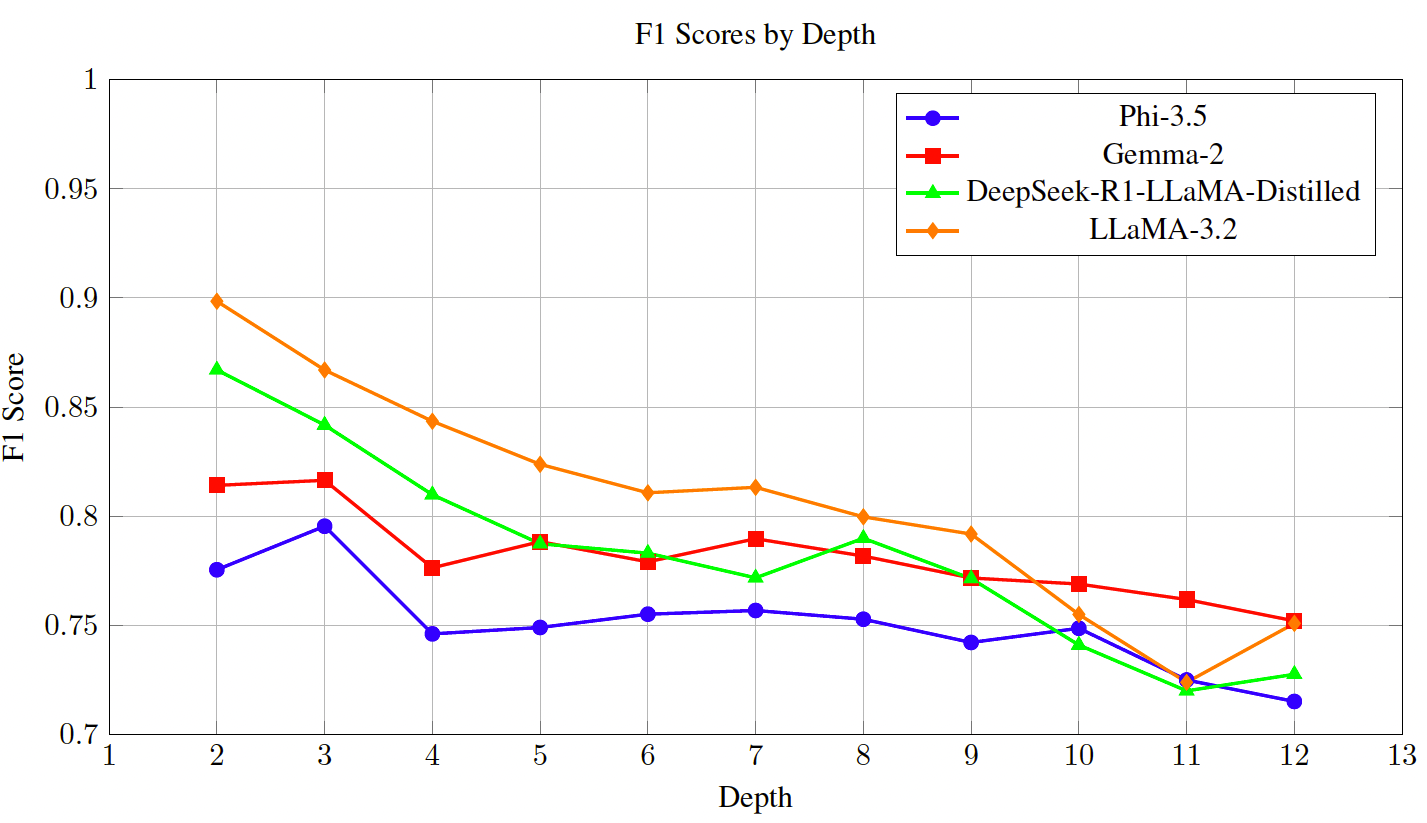}
    \caption{Consolidated F1 Scores [Silver MBSE AMR3.0 Dataset]}
    \label{fig:consolidated_f1_silver_plot}
\end{figure}

\begin{figure}[H]
    \centering
    \includegraphics[width=0.8\textwidth]{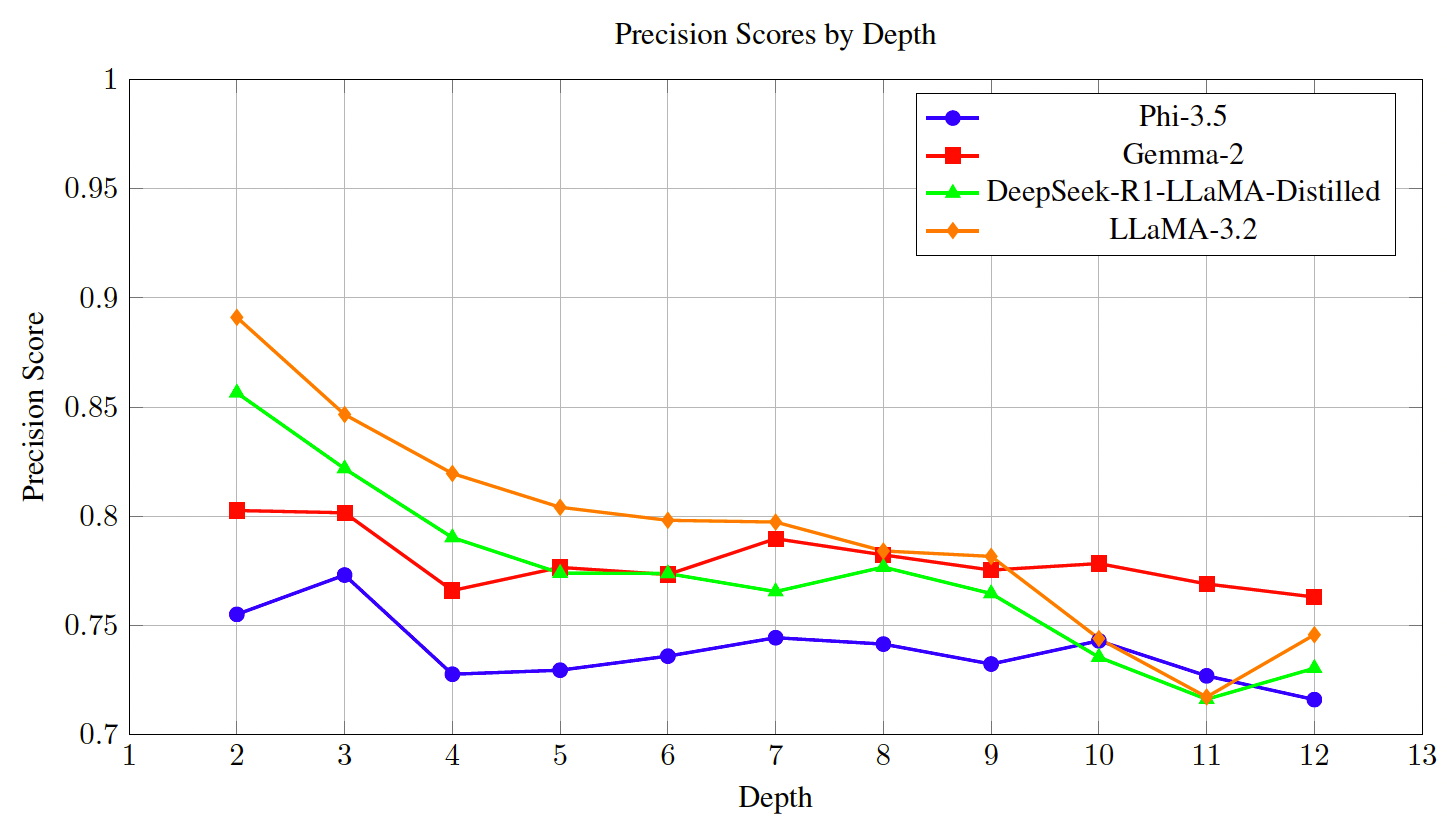}
    \caption{Consolidated Precision Scores [Silver MBSE AMR3.0 Dataset]}
    \label{fig:consolidated_precision_silver_plot}
\end{figure}

\begin{figure}[H]
    \centering
    \includegraphics[width=0.8\textwidth]{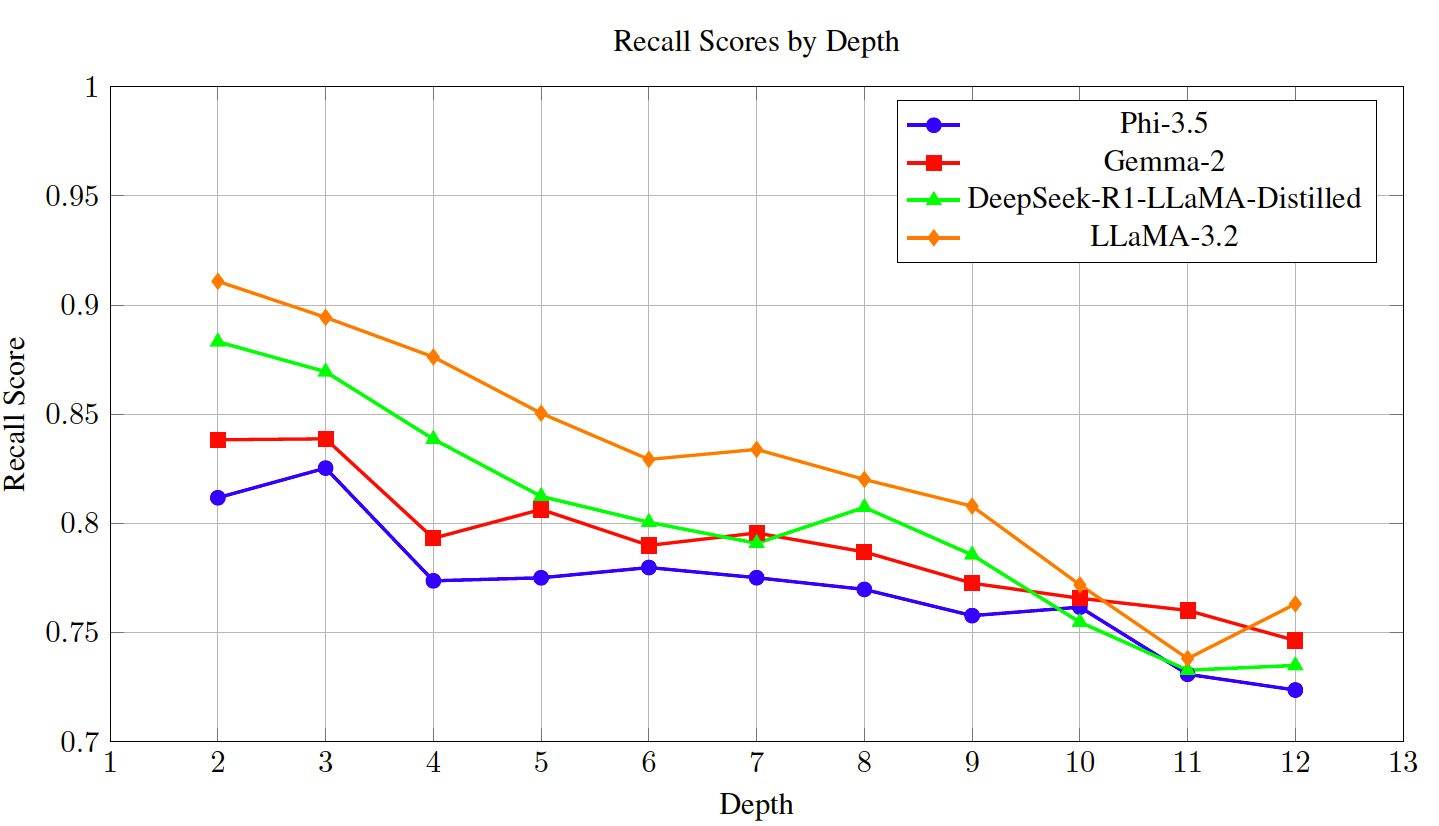}
    \caption{Consolidated Recall Scores [Silver MBSE AMR3.0 Dataset]}
    \label{fig:consolidated_recall_silver_plot}
\end{figure}

\subsubsection{Structural Validation across Depth Analysis}
As shown in Figure \ref{fig:consolidated_structural_silver}, our analysis of structural validation on Silver MBSE data reveals three key observations. 

Firstly, Phi-3.5, despite having a smaller parameter count (3.8B), demonstrates exceptional structural robustness across all depths, maintaining validity rates above 93\% even at depth 12 (466/500 valid).
    
Secondly, LLaMA-3.2 shows a distinct threshold limitation, maintaining strong structural validity through depth 6 (496/500 valid, 99.2\%) before experiencing exponential deterioration beyond depth 9 (475/500 valid, 95\%) to  depth 12 (382/500 valid, 76.4\%). This reinforces the smaller Gold AMR3.0 dataset findings in Section \ref{sec:depth-analysis-summarised}. This again indicates that LLaMA-3.2's language pre-trained nature may prioritise efficiency over structural coherence when handling extremely complex graphs.
    
Lastly, Gemma-2 shows the most severe structural degradation, with validity plummeting from 99.4\% at depth 2 (497/500) to merely 64\% at depth 12 (320/500). This aligns with \citet{springer2025overtrained}'s findings, as well as our findings in Section \ref{sec:semantic-validation-depth-analysis}, that Gemma-2 degrades with additional fine-tuning—while effective with simple structures, it fails to maintain coherence in complex hierarchical relationships as fine-tuning appears to compromise its generalisation capabilities.

\subsubsection{Semantic Validation across Depth Analysis}
As illustrated in Figures \ref{fig:consolidated_f1_silver_plot}, \ref{fig:consolidated_precision_silver_plot}, and \ref{fig:consolidated_recall_silver_plot}, our analysis of semantic validation on silver data reveals three key patterns. 

Firstly, all models exhibit a convergence phenomenon at the highest complexity levels (depths 11-12), with F1 scores narrowing to a range of just 0.03 (from 0.72 to 0.75) compared to a range of 0.13 at depth 2 (from 0.77 to 0.90). This convergence suggests that at extreme complexity levels, architectural and training differences become less significant, and all models approach a similar performance ceiling—potentially indicating a fundamental limit in current transformer-based approaches to handling highly complex semantic structures regardless of parameter count or training methodology.

Secondly, unlike the Gold AMR3.0 dataset where Precision consistently exceeded Recall, the Silver MBSE dataset shows Recall consistently exceeding Precision across all models. This pattern reversal suggests models prioritising comprehensive coverage (Recall) over exact concept matching (Precision) when evaluated on a larger sample size.
    
Lastly, LLaMA-3.2 demonstrates superior semantic performance at shallow depths (F1=0.90 at depth 2) but exhibits the steepest decline at deeper levels (F1=0.73 at depth 11, a 19\% drop). This contrasts with Phi-3.5, which shows the most consistent performance across all depths with minimal variation (F1=0.79 at depth 3 to F1=0.72 at depth 12, only a 9\% drop). This suggests that while LLaMA-3.2's extensive pretraining on diverse language corpora provides strong semantic understanding for simpler structures, this advantage diminishes with increasing graph complexity, whereas Phi-3.5's instruction-tuned approach yields more consistent performance across complexity levels.
\subsection{Subset Analysis}

To evaluate model robustness across different text domains, we conduct a comprehensive comparative analysis across all subsets of the LDC2020T02 Gold AMR 3.0 corpus. 

Table \ref{tab:dataset_distribution} shows the distribution of AMR graphs across different subsets, their abbreviations in bolded brackets, and their characteristics.
Table \ref{tab:model-subset-analysis} presents a comparative analysis of the four models across these six test subsets from LDC2020T02 Gold AMR3.0 dataset.
The graphs for this section are included in the Appendix in Section \ref{sec:appendix-subset-analysis-graphs}.

Due to the uneven distribution of AMR graphs of different depths within each subset dataset, we were unable to standardise the number of samples per depth category across datasets. This limitation is considered when interpreting cross-domain performance comparisons.

The comparative analysis helps identify potential domain biases in the model and informs strategies for improving cross-domain generalisation. The diverse nature of these subsets provides valuable insights into the model's performance across different linguistic contexts:

\begin{table}[ht]
\centering
\begin{tabular}{|l|r|p{8cm}|}
\hline
\textbf{Dataset} & \textbf{Size} & \textbf{Description} \\
\hline
BOLT DF MT [\textbf{Bolt}] & 133 & Discussion forum data with machine translation content \\
\hline
Weblog and WSJ [\textbf{Consensus}] & 100 & Mixed weblog entries and Wall Street Journal articles \\
\hline
BOLT DF English [\textbf{DFA}] & 229 & Native English discussion forum content \\
\hline
LORELEI [\textbf{Lorelei}] & 527 & Crisis and disaster scenarios with specialised vocabulary and complex causal relationships \\
\hline
Proxy Reports [\textbf{Proxy Reports}] & 823 & General news and formal report-style content with diverse topic coverage \\
\hline
Xinhua MT [\textbf{Xinhua MT}] & 86 & Translated Chinese news articles from Xinhua News Agency \\
\hline
\end{tabular}
\caption{Test Set Distribution and Characteristics}
\label{tab:dataset_distribution}
\end{table}

The F1, Precision, and Recall values are the mean and confidence intervals of the performance metrics across the 10 depth levels.
The Mean Error Count is the mean number of errors across the number of available depth levels within the subset.

The best subset scores for each metric are highlighted in bold.
The confidence intervals are calculated at the 95\% confidence level using the standard error of the mean across the depth levels, reflecting the variability in model performance across different graph complexities.

\begin{table}[H]
    \centering
    \resizebox{\linewidth}{!}{%
    \begin{tabular}{|l|c|c|c|c|c|c|c|}
    \hline
    \multirow{2}{*}{\textbf{Model}} & \multirow{2}{*}{\textbf{Metric}} & \multicolumn{6}{c|}{\textbf{Subset}} \\
    \cline{3-8}
     &  & Bolt & Lorelei & Consensus & DFA & Xinhua MT & Proxy Reports \\
    \hline
    \multirow{4}{*}{Phi-3.5} & F1 Score & 0.42 ± 0.08 & 0.38 ± 0.12 & 0.40 ± 0.06 & 0.37 ± 0.10 & 0.36 ± 0.08 & \textbf{0.52 ± 0.14} \\
    \cline{2-8}
     & Precision & 0.45 ± 0.08 & 0.40 ± 0.10 & 0.42 ± 0.05 & 0.39 ± 0.09 & 0.38 ± 0.07 & \textbf{0.55 ± 0.14} \\
    \cline{2-8}
     & Recall & 0.40 ± 0.09 & 0.36 ± 0.13 & 0.38 ± 0.07 & 0.35 ± 0.11 & 0.34 ± 0.09 & \textbf{0.49 ± 0.15} \\
    \cline{2-8}
     & Mean Error Count & 0.25 & \textbf{0.00} & 0.13 & 0.22 & 0.75 & 0.14 \\
    \hline
    \multirow{4}{*}{Gemma-2} & F1 Score & 0.40 ± 0.09 & 0.37 ± 0.07 & 0.35 ± 0.05 & 0.34 ± 0.10 & 0.35 ± 0.08 & \textbf{0.45 ± 0.20} \\
    \cline{2-8}
     & Precision & 0.42 ± 0.08 & 0.39 ± 0.06 & 0.37 ± 0.04 & 0.36 ± 0.09 & 0.38 ± 0.07 & \textbf{0.48 ± 0.19} \\
    \cline{2-8}
     & Recall & 0.38 ± 0.10 & 0.35 ± 0.08 & 0.33 ± 0.06 & 0.32 ± 0.11 & 0.33 ± 0.09 & \textbf{0.42 ± 0.21} \\
    \cline{2-8}
     & Mean Error Count & 1.25 & 2.13 & 1.38 & \textbf{1.22} & 1.50 & 2.00 \\
    \hline
    \multirow{4}{*}{DeepSeek-R1-LLaMA-Distilled} & F1 Score & 0.35 ± 0.05 & 0.30 ± 0.14 & 0.35 ± 0.05 & 0.32 ± 0.08 & 0.38 ± 0.10 & \textbf{0.40 ± 0.24} \\
    \cline{2-8}
     & Precision & 0.37 ± 0.04 & 0.32 ± 0.13 & 0.37 ± 0.04 & 0.34 ± 0.07 & 0.40 ± 0.09 & \textbf{0.43 ± 0.23} \\
    \cline{2-8}
     & Recall & 0.33 ± 0.06 & 0.28 ± 0.15 & 0.33 ± 0.06 & 0.30 ± 0.09 & \textbf{0.36 ± 0.11} & 0.37 ± 0.25 \\
    \cline{2-8}
     & Mean Error Count & 0.25 & \textbf{0.00} & 0.13 & 0.11 & 0.25 & \textbf{0.00} \\
    \hline
    \multirow{4}{*}{LLaMA-3.2} & F1 Score & 0.41 ± 0.07 & 0.32 ± 0.09 & 0.35 ± 0.05 & 0.38 ± 0.06 & 0.38 ± 0.10 & \textbf{0.50 ± 0.15} \\
    \cline{2-8}
     & Precision & 0.43 ± 0.06 & 0.34 ± 0.08 & 0.37 ± 0.04 & 0.40 ± 0.05 & 0.40 ± 0.09 & \textbf{0.53 ± 0.14} \\
    \cline{2-8}
     & Recall & 0.39 ± 0.08 & 0.30 ± 0.10 & 0.33 ± 0.06 & 0.36 ± 0.07 & 0.36 ± 0.11 & \textbf{0.47 ± 0.16} \\
    \cline{2-8}
     & Mean Error Count & 0.38 & 0.50 & \textbf{0.25} & 0.33 & \textbf{0.25} & 0.43 \\
    \hline
    \end{tabular}
    }
    \caption{Individual Model Performance Analysis Across Subsets}
    \label{tab:model-subset-analysis}
\end{table}

\subsubsection{Summarised Analysis}
\textbf{Semantic Validation}: All models demonstrate domain-specific semantic strengths, with Proxy Reports consistently yielding the highest F1 scores (0.40-0.52) and Lorelei data presenting the greatest challenge (0.28-0.40) as shown in Table \ref{tab:model-subset-analysis}. DeepSeek-R1-LLaMA-Distilled exhibits the most consistent cross-domain performance with the smallest performance gap (0.10) between its strongest and weakest domains. Interestingly, while models achieve higher mean performance on formal content (Proxy Reports), their consistency, as shown by narrower confidence intervals in Table \ref{tab:model-subset-analysis}, is actually greater on discussion forum and standard text data (Bolt, Consensus).

\textbf{Structural Validation}: DeepSeek-R1-LLaMA-Distilled and Phi-3.5 demonstrate superior structural robustness with perfect validity on specific domains (Lorelei and Proxy Reports for DeepSeek-R1-LLaMA-Distilled, Lorelei for Phi-3.5) as shown in Table \ref{tab:model-subset-analysis}. A striking inverse relationship exists between structural and semantic performance across domains, with models achieving the highest structural validity often showing lower semantic scores in the same domain. Gemma-2 exhibits significant structural challenges across all domains (1.22-2.13 errors per depth), aligning with findings that it is historically difficult to fine-tune.

\subsubsection{Structural Validity across Depth Analysis}
Our analysis of structural validity across different subsets in Section \ref{sec:appendix-subset-analysis-graphs} reveals three key observations. 

Firstly, DeepSeek-R1-LLaMA-Distilled and LLaMA-3.2, despite sharing architectural foundations, exhibit markedly different structural validity patterns across domains: DeepSeek-R1-LLaMA-Distilled achieves perfect validity (0.00 errors) on specialised Lorelei data and diverse Proxy Reports, while LLaMA-3.2 performs best on Consensus and Xinhua MT (both 0.25 errors) as shown in Table \ref{tab:model-subset-analysis}. This domain-specific divergence suggests that while both models leverage similar transformer architectures, DeepSeek-R1-LLaMA-Distilled's additional training optimisations (particularly its Chain-of-Thought capabilities) provide superior structural robustness for complex specialised content, while LLaMA-3.2's extensive pretraining excels with standard and translated content.

Secondly, a striking inverse relationship exists between structural and semantic performance across domains: models achieving the highest structural validity often show lower semantic scores in the same domain. For instance, DeepSeek-R1-LLaMA-Distilled achieves perfect structural validity on Lorelei (0.00 errors) but its lowest F1 score (0.30) in this domain, while showing higher semantic performance on Xinhua MT (F1: 0.38) despite more structural errors (0.25). Similarly, Phi-3.5 demonstrates perfect structural validity on Lorelei (0.00 errors) with moderate semantic performance (F1: 0.38), but achieves its highest semantic scores on Proxy Reports (F1: 0.52) with slightly more structural errors (0.14). This pattern suggests a fundamental trade-off between structural coherence and semantic richness that persists across model architectures and domains.

Lastly, Gemma-2 exhibits consistent structural challenges across all domains, with error counts significantly higher than other models (1.22-2.13 errors per depth). This, again, aligns with findings in Sections \ref{sec:depth-analysis-summarised} and Section \ref{sec:silver-analysis-summarised}, from \citet{springer2025overtrained} that Gemma-2 is historically difficult to fine-tune, as the model appears to struggle with maintaining structural coherence across specialised domains, suggesting its pre-training approach may compromise its ability to adapt to the structural requirements of AMR parsing.

\subsubsection{Semantic Validation across Depth Analysis}
As illustrated in the F1, Precision, and Recall plots for each model in Section \ref{sec:appendix-subset-analysis-graphs}, our analysis of semantic validation across different subsets reveals three key patterns. 

Firstly, all models demonstrate domain-specific semantic strengths, with Proxy Reports consistently yielding the highest F1 scores (0.40-0.52) and Lorelei data presenting the greatest challenge (0.30-0.38) as shown in Table \ref{tab:model-subset-analysis}. This consistent pattern across architecturally diverse models suggests that formal, well-structured content is inherently more amenable to semantic parsing than specialised crisis terminology, regardless of model design or training methodology.
    
Secondly, the confidence intervals reveal distinct model-specific uncertainty patterns across domains: Phi-3.5 and LLaMA-3.2 show the narrowest confidence intervals on Consensus (±0.04-0.07) and Bolt (±0.04-0.10) data, while all models exhibit substantially wider intervals on Proxy Reports (±0.14-0.24) as shown in Table \ref{tab:model-subset-analysis}. This suggests that while models may achieve higher mean performance on formal content, their consistency is actually greater on discussion forum and standard text. DeepSeek-R1-LLaMA-Distilled shows particularly high variance on Lorelei data (±0.14) and Proxy Reports (±0.24), indicating that its performance on these domains is less predictable—a critical consideration for applications requiring consistent results across similar inputs.

Lastly, DeepSeek-R1-LLaMA-Distilled exhibits the most consistent cross-domain performance with the smallest performance gap (0.10) between its strongest and weakest domains. This versatility could be attributed to its Chain-of-Thought capabilities, which provides robust semantic understanding across varying content types and enables effective extraction of semantic relationships in different domains.

\section{Conclusion}
\subsection{Discussion}
In this work, we have finetuned, evaluated, and analysed the performance of four open-source LLMs: LLaMA-3.2, DeepSeek-R1-LLaMA-Distilled, Gemma-2, and Phi-3.5 for AMR parsing. 

Our comprehensive evaluation across multiple dimensions has yielded several key insights. 
Using the LDC2020T02 Gold AMR 3.0 dataset, we have found that LLaMA-3.2 demonstrated superior semantic performance on the test split with an F1 score of 0.804, comparable to specialised SOTA parsers like APT + Silver (IBM). 

In our Depth Analysis, LLaMA 3.2's superior semantic performance (F1: 0.87) was further validated, while Phi-3.5 exhibited the most consistent structural validity, with a Mean Error Count of 0.3 errors per depth. 

These conclusions were validated at scale using the extensive Silver MBSE dataset, with an interesting observation that all models converged in F1 performance at higher complexity levels (depths 11-12). 

The Subset Analysis across different text domains in the LDC2020T02 Gold AMR 3.0 dataset highlighted domain-specific strengths, with Proxy Reports yielding the highest F1 scores (0.40-0.52) across all models, and DeepSeek-R1-LLaMA-Distilled exhibiting the most consistent cross-domain performance. 

These findings demonstrate that LLMs can effectively tackle specialised tasks like AMR parsing with minimal architectural modifications, approaching the performance of dedicated complex parsers.

\subsection{Limitations}
The limitations of this work is mostly hardware related. 
Given more computational resources, we could have explored finetuning larger sized models and compared the semantics and structural performance differences between model sizes of the same model. 

\subsection{Future Work}
Future work could explore several promising directions. 
First, we can explore the efficacy of finetuning these models using the Silver MBSE dataset to further improve their performance. 
Secondly, we can explore finetuning of newer versions of these open-source models, which are likely to be released in the future. 
Lastly, we can even explore the efficacy of closed-source current SOTA reasoning models like OpenAI's GPT-o4 and Anthropic's Claude Sonnet 4 for AMR parsing. 

\section{Appendix}
\label{sec:appendix}

\subsection{Training and Evaluation Loss Figures}
\label{sec:appendix-training-evaluation-loss}

\subsection{Subset Analysis Graphs}
\label{sec:appendix-subset-analysis-graphs}

\begin{figure}[H]
    \centering
    \includegraphics[width=0.8\textwidth]{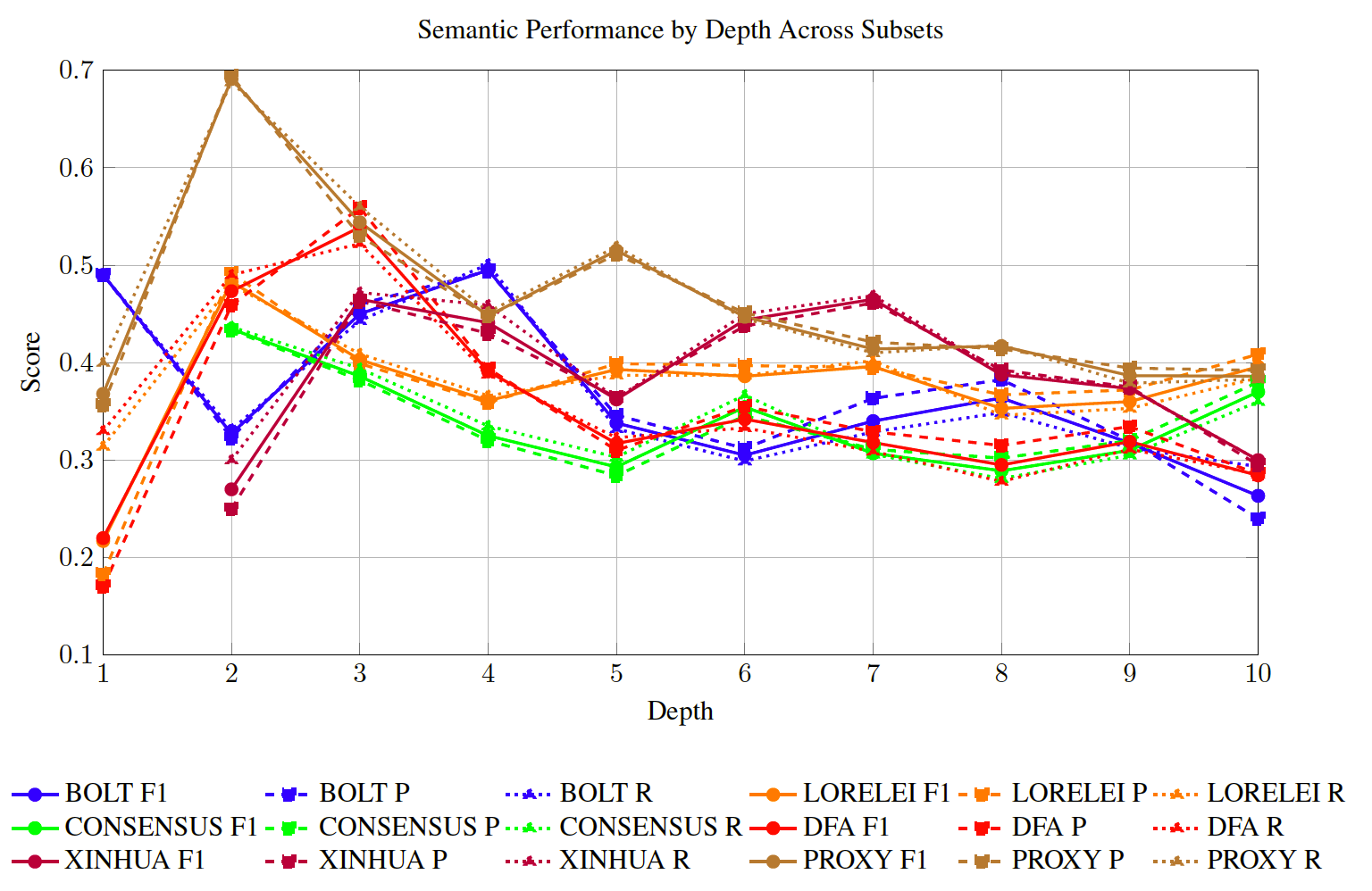}
    \caption{Semantic Performance Across Subsets [Phi-3.5]}
\end{figure}

\begin{figure}[H]
    \centering
    \includegraphics[width=0.8\textwidth]{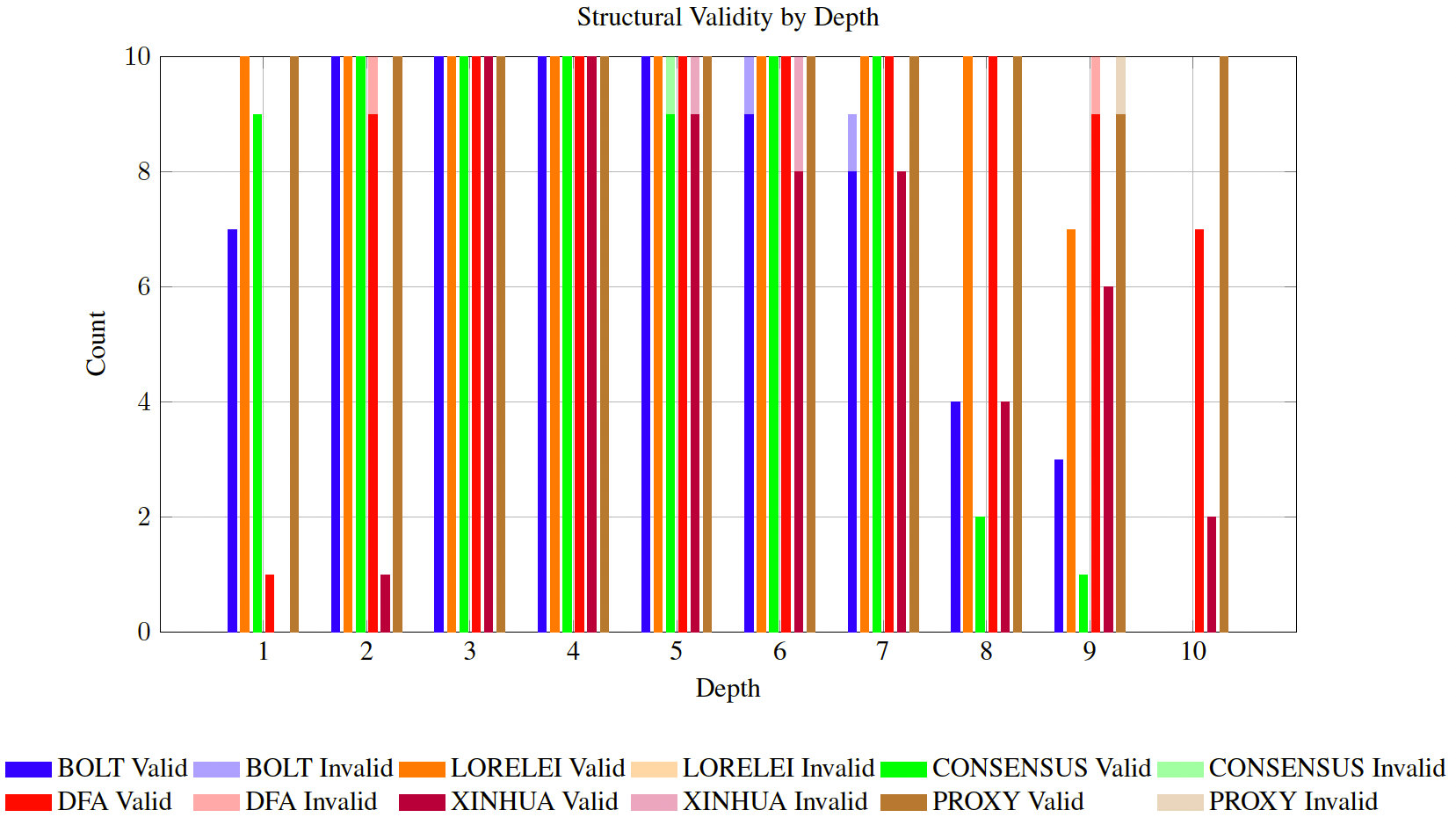}
    \caption{Structural Performance Across Subsets [Phi-3.5]}
\end{figure}

\begin{figure}[H]
    \centering
    \includegraphics[width=0.8\textwidth]{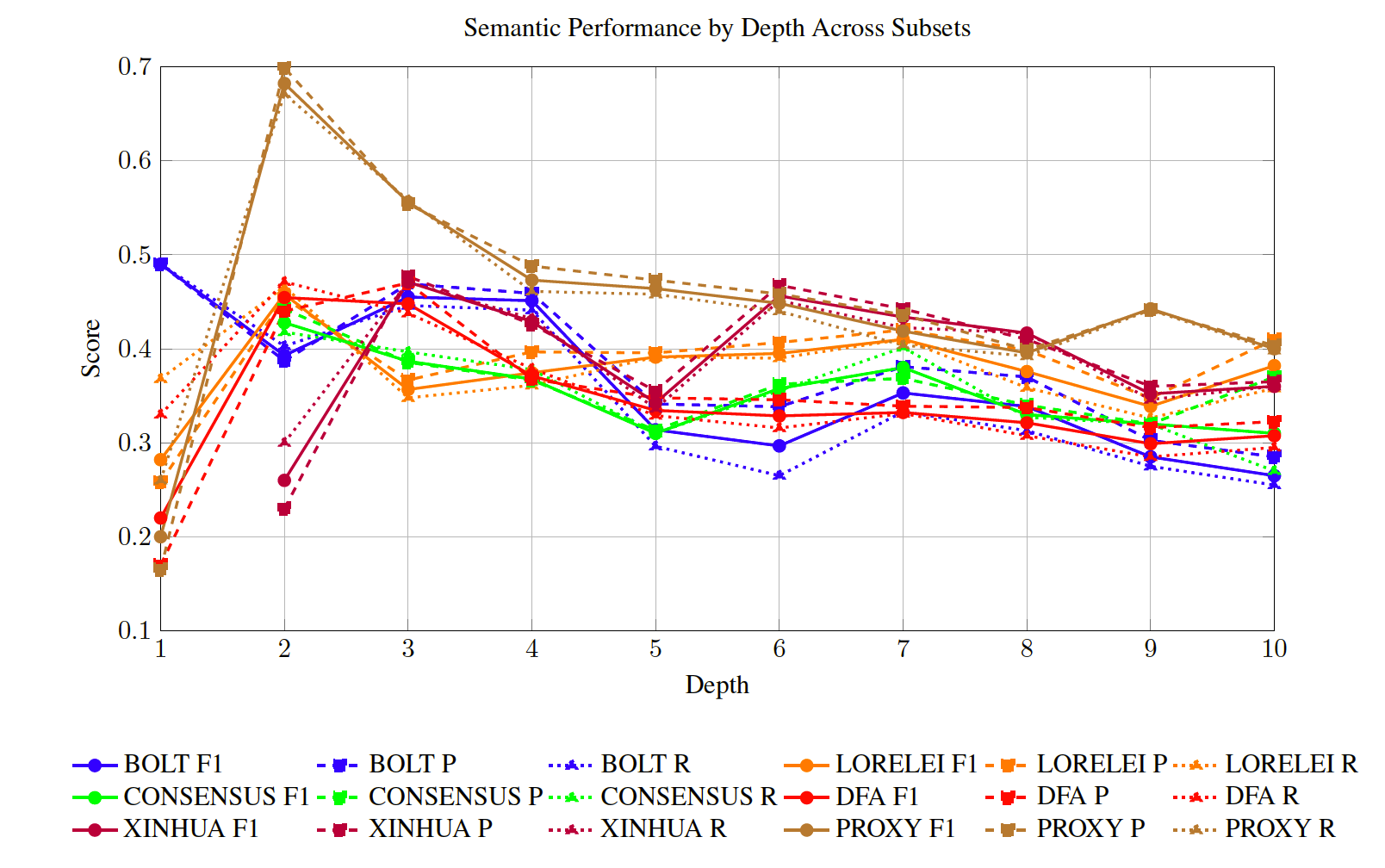}
    \caption{Semantic Performance Across Subsets [Gemma-2]}
\end{figure}

\begin{figure}[H]
    \centering
    \includegraphics[width=0.8\textwidth]{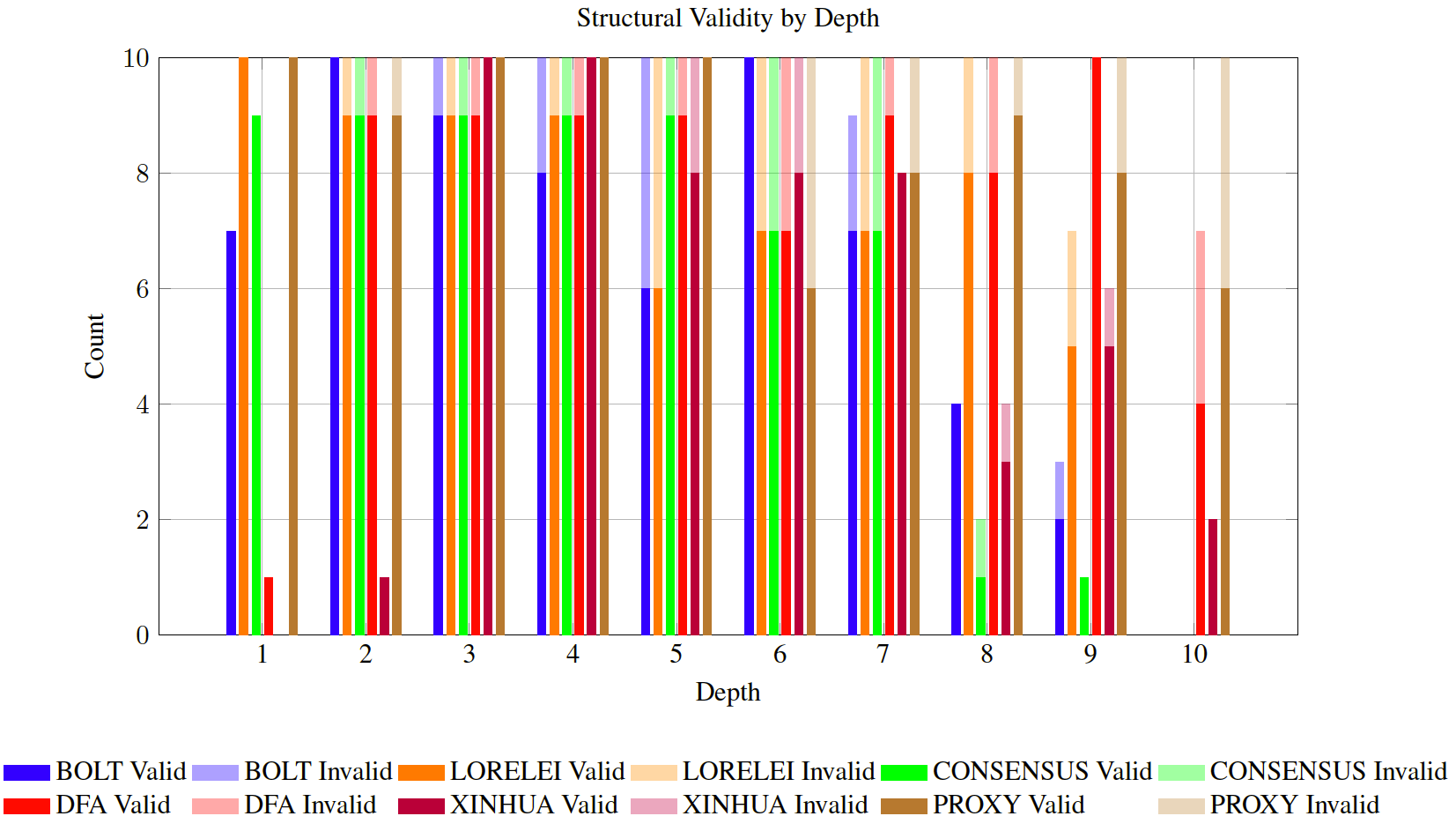}
    \caption{Structural Performance Across Subsets [Gemma-2]}
\end{figure}

\begin{figure}[H]
    \centering
    \includegraphics[width=0.8\textwidth]{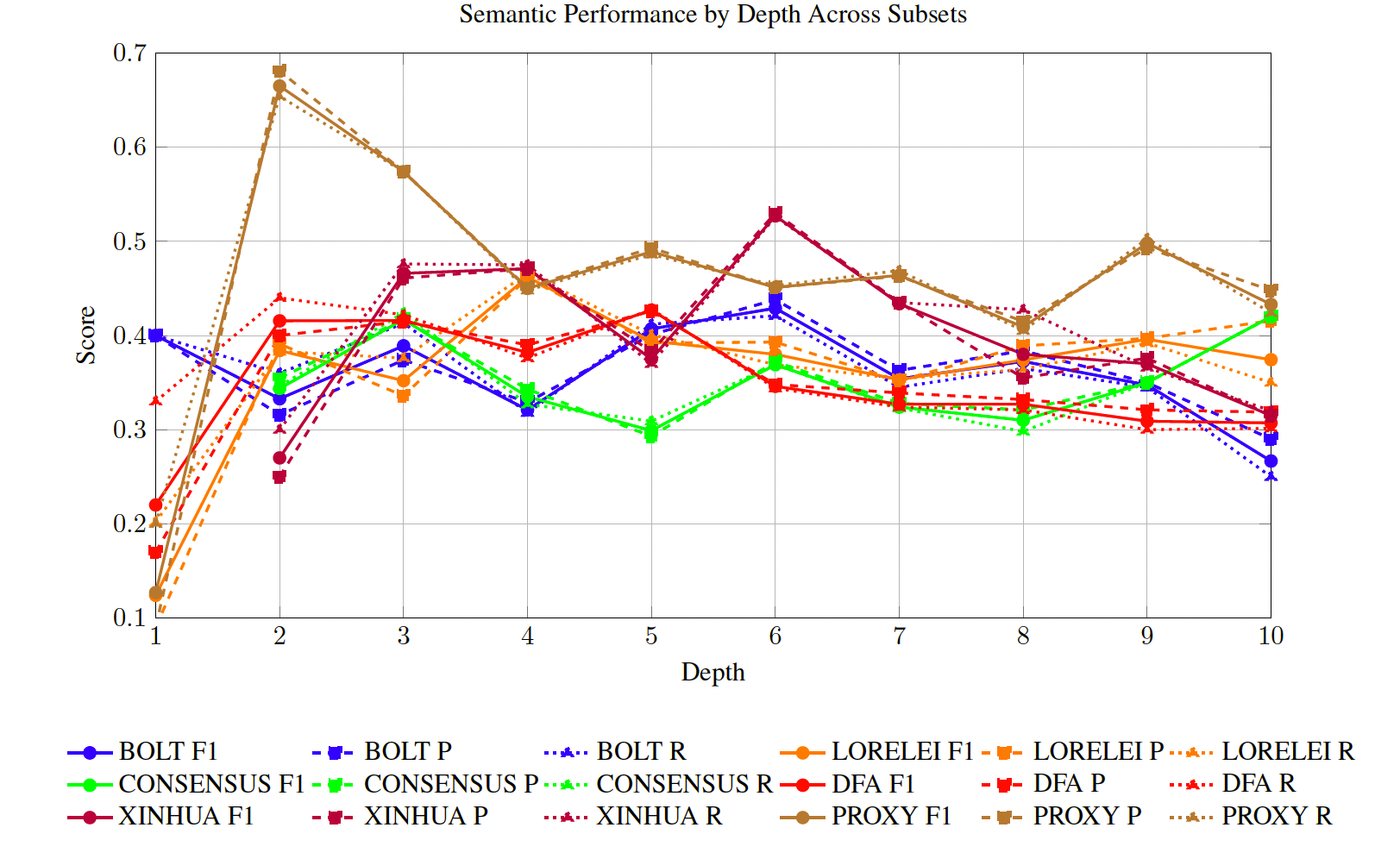}
    \caption{Semantic Performance Across Subsets [DeepSeek-R1-LLaMA-Distilled]}
\end{figure}

\begin{figure}[H]
    \centering
    \includegraphics[width=0.8\textwidth]{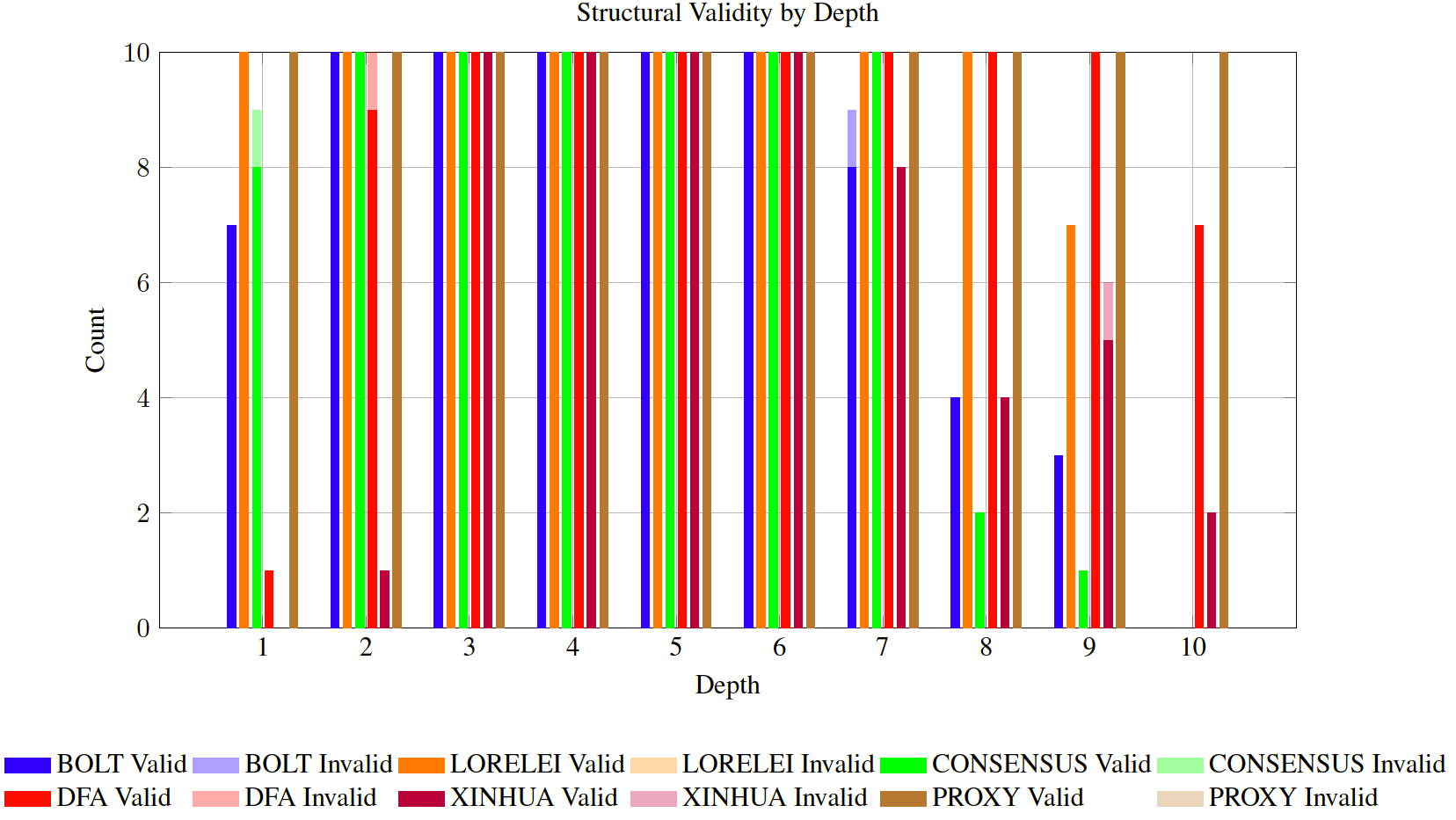}
    \caption{Structural Performance Across Subsets [DeepSeek-R1-LLaMA-Distilled]}
\end{figure}

\begin{figure}[H]
    \centering
    \includegraphics[width=0.8\textwidth]{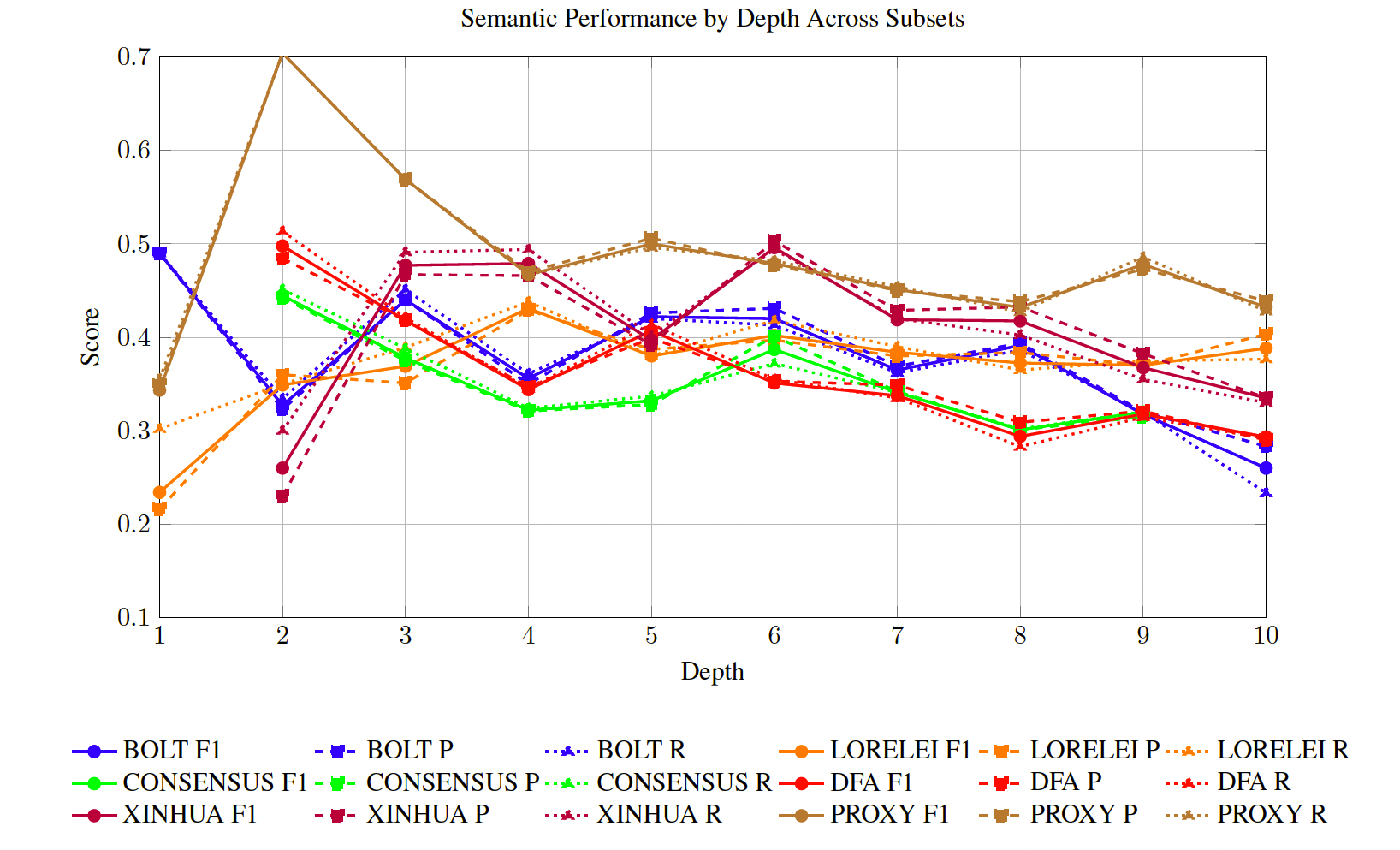}
    \caption{Semantic Performance Across Subsets [LLaMA-3.2]}
\end{figure}

\begin{figure}[H]
    \centering
    \includegraphics[width=0.8\textwidth]{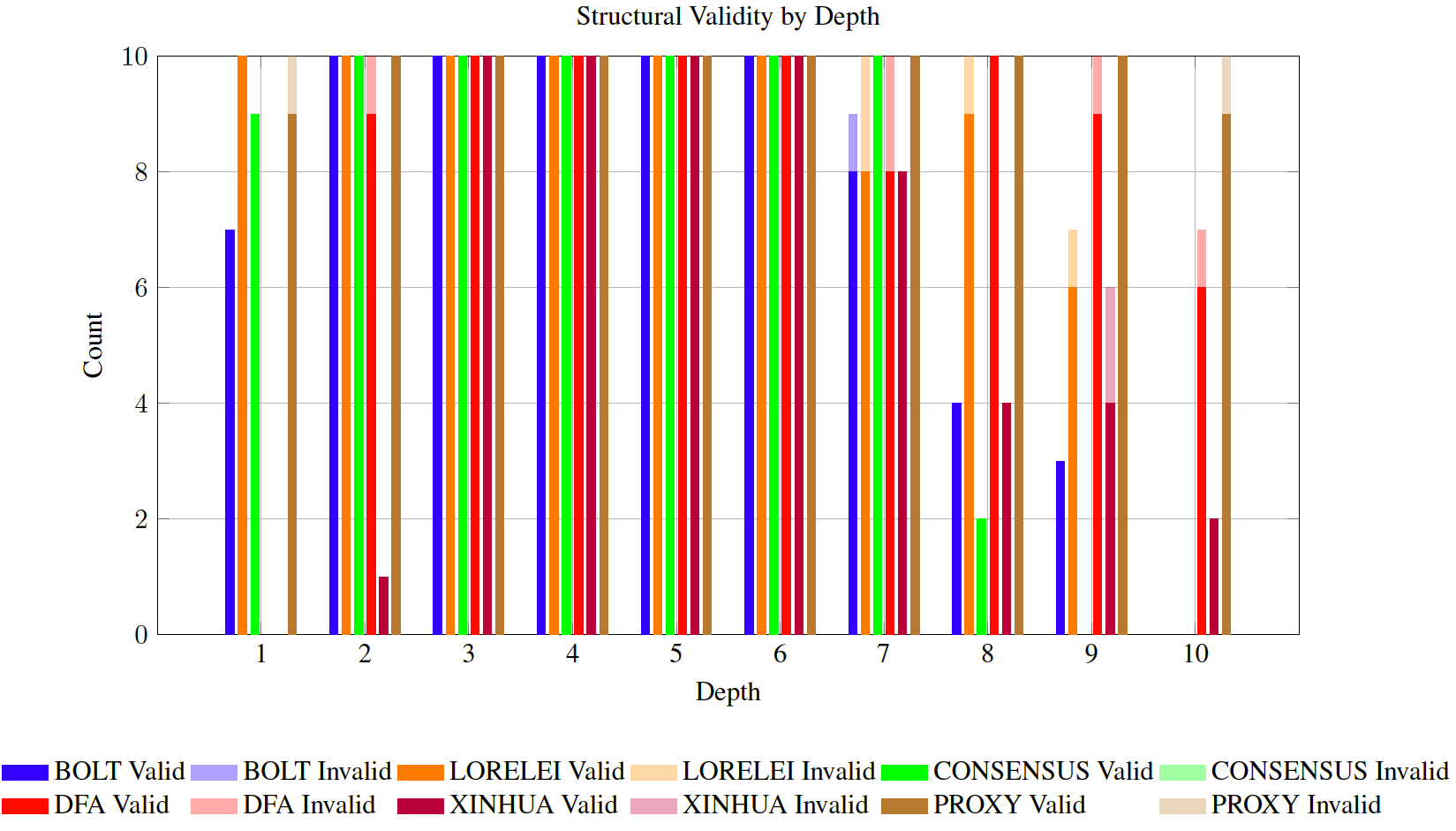}
    \caption{Structural Performance Across Subsets [LLaMA-3.2]}
\end{figure}

\bibliographystyle{plainnat}
\bibliography{references}
\end{document}